\documentclass{article} 
\usepackage{iclr2027_conference,times}

\usepackage{amsmath,amsfonts,bm}

\def\eqref#1{equation~\ref{#1}}

\def\1{\bm{1}}

\DeclareMathAlphabet{\mathsfit}{\encodingdefault}{\sfdefault}{m}{sl}
\SetMathAlphabet{\mathsfit}{bold}{\encodingdefault}{\sfdefault}{bx}{n}

\DeclareMathOperator*{\minimize}{Minimize}

\usepackage{xcolor}
\usepackage[most]{tcolorbox}
\usepackage[framemethod=tikz]{mdframed}
\usepackage{xparse}
\usepackage{tikz}

\newtcbox{\bracketeq}{
  nobeforeafter, enhanced, colback=gray!10, frame empty, math upper,
  boxsep=2pt, left=8pt, right=8pt, top=4pt, bottom=4pt, arc=4pt,
  tcbox raise base, 
  overlay={
    \def\thk{1.0pt}   
    \def\stub{4pt}    
    \draw[line width=\thk] (frame.north west) -- (frame.south west);
    \draw[line width=\thk] (frame.north east) -- (frame.south east);
    \draw[line width=\thk] (frame.north west) -- ([xshift=\stub]frame.north west);
    \draw[line width=\thk] (frame.south west) -- ([xshift=\stub]frame.south west);
    \draw[line width=\thk] (frame.north east) -- ([xshift=-\stub]frame.north east);
    \draw[line width=\thk] (frame.south east) -- ([xshift=-\stub]frame.south east);
  }
}
\newtcbox{\boxeq}{%
  enhanced,
  on line,
  nobeforeafter,
  math upper,          
  frame empty,
  colback=gray!10,
  boxsep=2pt, left=4pt, right=4pt, top=4pt, bottom=4pt,
  arc=3pt,
  tcbox raise base,
  overlay={
    \def\r{3pt}         
    \def\thk{1.0pt}     
    \def\stub{4pt}      
    \def\wipe{1pt}      
    \draw[line width=\thk, rounded corners=\r]
      (frame.north west) rectangle (frame.south east);
    \fill[white]
      ([xshift=\stub,yshift=\wipe]frame.north west)
      rectangle
      ([xshift=-\stub,yshift=-\wipe]frame.north east);
    \fill[white]
      ([xshift=\stub,yshift=\wipe]frame.south west)
      rectangle
      ([xshift=-\stub,yshift=-\wipe]frame.south east);
  }
}

\usepackage{hyperref}
\usepackage{url}
\usepackage{subcaption}
\usepackage{booktabs} 
\usepackage{colortbl}
\definecolor{tableheader}{HTML}{DDDDDD}
\definecolor{tableours}{HTML}{DDDDDD}
\definecolor{tableours2}{HTML}{f4f4f4}
\definecolor{tablegroup}{HTML}{F5F7F9}
\usepackage{mdframed}
\usepackage{xcolor}
\usepackage{paralist}
\usepackage{enumitem}
\usepackage{wrapfig}
\usepackage{placeins}
\usepackage{adjustbox}
\usepackage{algpseudocode}
\usepackage{algorithm}
\usepackage{multirow}
\usepackage{multicol}
\usepackage{array, makecell}
\usepackage{mathtools}
\usepackage{amsthm}
\newtheorem{proposition}{Proposition}
\newtheorem{corollary}[proposition]{Corollary}
\newcommand{\safeincludegraphics}[2][]{%
 \IfFileExists{#2}{%
 \includegraphics[#1]{#2}%
 }{%
 \fbox{%
  \parbox[c][0.22\textheight][c]{0.9\linewidth}{%
  \centering
  Missing figure asset%
  }%
 }%
 }%
}
\usepackage{float}
\newfloat{model}{thp}{lop}
\floatname{model}{Model}

\usepackage{multirow,mdframed}
\mdfsetup{skipabove=0pt,skipbelow=0pt}
\mdfdefinestyle{model}{%
 innertopmargin=0pt,
 innerbottommargin=0pt,
 innerleftmargin=0pt,
 innerrightmargin=0pt,
 skipbelow=0pt,%
 skipabove=0pt,
 splitbottomskip=0pt,
 splittopskip=0pt,
 leftmargin =0pt,%
 rightmargin=0pt,
 splittopskip=0pt,
 usetwoside=false,
}

\usepackage{hyperref}
\usepackage{url}
\usepackage[normalem]{ulem} 

\hypersetup{
  colorlinks=true,
  allcolors=blue,
  pdfborder={0 0 0}
}

\title{Constrained Graph Diffusion for \\ Mixed-Integer Optimization}

\author{%
\normalfont
\begin{tabular}{@{}c@{}}
\noalign{\vskip 0.8em}
  \begin{tabular*}{\textwidth}{@{\extracolsep{\fill}}ccc@{}}
    \textbf{Vincenzo Di Vito}\textsuperscript{1} &
    \textbf{Yusuf Guven}\textsuperscript{1} &
    \textbf{Babak Badnava}\textsuperscript{2}
    \\[0.4em]
    \textbf{Deepjyoti Deka}\textsuperscript{3} &
    \textbf{Kaarthik Sundar}\textsuperscript{4} &
    \textbf{Ferdinando Fioretto}\textsuperscript{1}
  \end{tabular*}
  \\[1.6em]
  {\normalfont\small
    \textsuperscript{1}University of Virginia
    \qquad
    \textsuperscript{2}Siemens Energy}
    \qquad
    \textsuperscript{3}MIT
    \qquad
    \textsuperscript{4}Los Alamos National Laboratory
  \\[1.6em]
  {\normalfont\footnotesize
    \textsuperscript{1}\texttt{\{eda8pc,ngu6fh,fioretto\}@virginia.edu}}
  \\[0.2em]
  {\normalfont\footnotesize
    \textsuperscript{2}\texttt{babak.badnava@siemens-energy.com}}
  \\[0.2em]
  {\normalfont\footnotesize
    \textsuperscript{3}\texttt{deepj87@mit.edu}
    \qquad
    \textsuperscript{4}\texttt{kaarthik@lanl.gov}}
\end{tabular}%
}

\iclrfinalcopy 
\begin{document}

\maketitle

\begin{abstract}
This paper proposes a novel learning-based approach to approximately solve instances of mixed-integer optimization problems. These problems are computationally challenging, as they require jointly determining discrete and continuous decisions while satisfying complex combinatorial constraints. 
We introduce Constrained Graph Diffusion
(CGD), a learning-based framework that approximately solves recurring
instances of such problems by learning a conditional distribution over their
discrete decisions. CGD uses a graph-based diffusion model and incorporates
constraint information directly into the reverse diffusion process, steering
intermediate predictions toward the feasible region throughout generation.
By operating on continuous relaxations of the discrete variables, CGD defines
a differentiable constrained generation pathway up to terminal discrete
recovery. Once the discrete decision is recovered and fixed, a numerical optimizer solves the remaining continuous problem, avoiding online combinatorial search over the binary variables while retaining numerical optimization for continuous completion.
We evaluate CGD on AC-OPF with branch switching and discrete portfolio optimization, demonstrating substantial improvements in feasibility and solution quality over learning-based baselines while achieving speedups of up to $543\times$ over state-of-the-art MIP solvers on large instances.
\end{abstract}



%

\raggedbottom
\sloppy

\section{Introduction}

Mixed-integer optimization provides a powerful framework for modeling problems involving both discrete and continuous variables. Applications arise in a wide range of domains, including power systems~\citep{hedman2010cooptimization}, logistics~\citep{amiri2006distribution} and finance~\citep{bienstock1996computational}. 

This work considers problems of the form
\begin{equation}
\begin{aligned}
\minimize_{x,z} \;\; f(x,z)\quad
\text{s.t.}\quad & g(x,z)\le 0,\\
& x\in\mathcal X\subseteq\mathbb R^n, \;
& z\in\{0,1\}^{m},
\end{aligned}
\label{eq:mip}
\end{equation}
where $x\!\in\!\mathcal X$ are the \emph{continuous} decision variables, $z$ the \emph{binary} decision variables, $f$ is the objective, and $g$ collects the problem constraints, with equality constraints represented as pairs of inequalities. 
In the most general case, Problem~(\ref{eq:mip}) is a nonconvex mixed-integer nonlinear program. In such problems, two distinct sources of difficulty compound: \emph{(1) Combinatorial structure:} the binary variables $z$ induce a search space of size $2^{m}$, forcing exact methods to branch over an exponential number of assignments; \emph{(2) Nonconvexity:} the objective $f$ and the constraints $g$ are nonconvex, so even a single fixed 
 relaxation is a nonconvex program with no guarantee of global optimality.

When instances of Problem~(\ref{eq:mip}) share the same structure, Learning to Optimize can amortize solution effort across instances~\citep{bengio2021machine,kotary2021end}. Existing approaches either improve decisions within conventional solvers~\citep{khalil2016learning,gasse2019exact,nair2020solving} or directly predict solutions through learned surrogates and differentiable optimization layers~\citep{wilder2018melding,ferbermipaal}. The former retain online combinatorial search, whereas the latter may overlook multiple near-optimal assignments and produce fractional or infeasible decisions.

The proposed approach learns instead a conditional distribution over the discrete variables. Once an assignment $\hat z$ is fixed, a numerical optimizer recovers the continuous variables $x$, avoiding online combinatorial search without eliminating the potential nonconvexity of the continuous subproblem. Diffusion models naturally represent such multimodal distributions and have been applied to graph-structured problems~\citep{sun2023difusco,sanokowski2024diffusion}. However, existing diffusion solvers do not natively enforce combinatorial constraints during generation, although an invalid discrete assignment may render the downstream continuous problem infeasible, as shown in Fig.~\ref{fig:infeasible_topology_example}.

\begin{figure}[t]
    \centering
    \includegraphics[width=.95\linewidth]{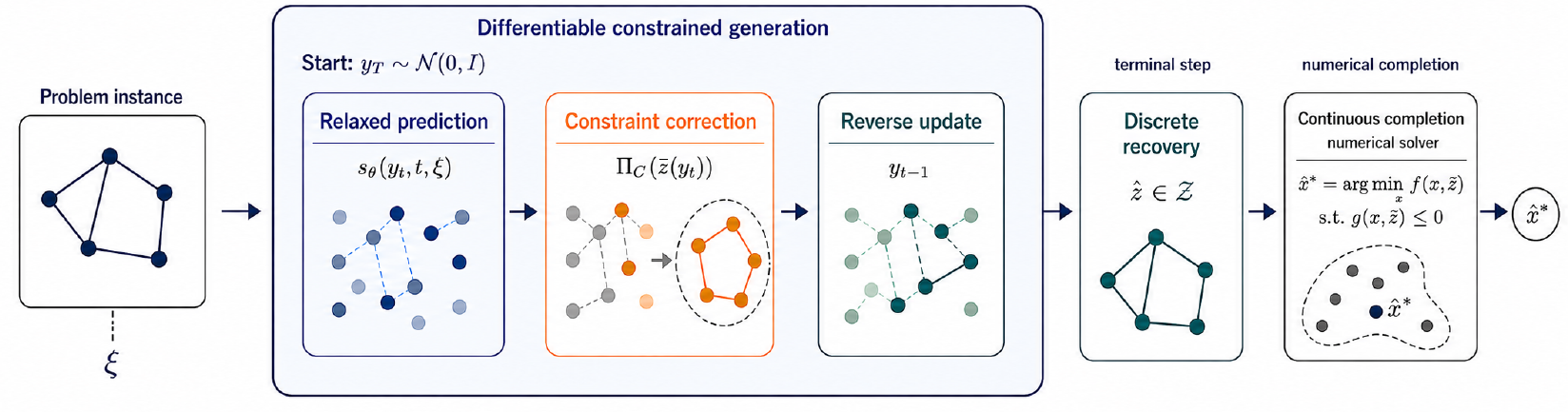}
    \caption{
Overview of the generation phase of CGD. At each reverse step $t$, the denoiser produces a relaxed binary decision, whose Monte Carlo estimate $\bar z_t$ is corrected with respect to the relaxed feasible set
$\mathcal C_\xi$. The resulting correction is incorporated into the reverse
update. Terminal recovery produces the binary decision $\hat z$, while a
numerical solver recovers the continuous variables $\hat x$.
}
    \label{fig:overview_framework}
\end{figure}

Motivated by this gap, we introduce \emph{Constrained Graph Diffusion} (CGD), a diffusion framework that incorporates feasibility corrections directly into the reverse denoising process. CGD operates in a differentiable fashion over relaxed binary decisions, allowing each correction to influence all subsequent denoising steps and steer generation toward the feasible set. Hard discretization is then deferred to the terminal step, after which the recovered binary decision is fixed and the remaining continuous variables are obtained using a standard numerical solver, ensuring validity (Fig.~\ref{fig:overview_framework}).

\textbf{Contributions.} The paper makes the following contributions:
{\bf (1)} It introduces Constrained Graph Diffusion (CGD), a novel end-to-end differentiable generation framework over relaxed discrete decisions. It uses application-specific feasibility corrections to steer the reverse trajectory, with hard recovery deferred to the terminal step.
{\bf (2)} It proposes a learning-assisted decomposition that generates the discrete variables and recovers the continuous variables through a downstream numerical optimization problem, thereby avoiding online combinatorial search, while retaining solution validity. 
{\bf (3)} It evaluates CGD on AC optimal power flow with branch switching and mixed-integer portfolio optimization. Across these highly nonlinear problems, CGD improves feasibility and solution quality over learning-based baselines while achieving substantial speedups over MIP solvers.



\section{Related Work}
\label{sec:related_work}
Our work lies at the intersection of learning for mixed-integer optimization and diffusion models for combinatorial optimization. We summarize the closest connections here and provide a comprehensive methodological review in Appendix~\ref{app:related_work}.

\textbf{Learning for Mixed-Integer Optimization.}
Learning-based methods can accelerate repeated optimization by predicting solutions from instance data~\citep{bengio2021machine,kotary2021end}. One line of work learns components of conventional solvers, including branching, primal heuristics, and cut selection~\citep{khalil2016learning,gasse2019exact,nair2020solving,tang2020learningcut,kannan2025strong}. These methods preserve the solver's feasibility and certification mechanisms but retain online combinatorial search. More direct approaches predict solutions or optimization strategies~\citep{Bertsimas_learn_19,wilder2018melding,ferbermipaal,tang2025learningoptimizemixedintegernonlinear}; however, they commonly return a single assignment or rely on relaxations, rounding, repair, or downstream optimization. CGD instead learns a distribution over discrete variables, avoiding online branch-and-bound while retaining numerical optimization for continuous completion.
\\
\textbf{Diffusion Models for Combinatorial Optimization.}
Diffusion models have produced high-quality solutions for graph-structured combinatorial problems~\citep{sun2023difusco,sanokowski2024diffusion,hong2024unsupervised,feng2026cam,lei2025diffuada}. Existing approaches typically address constraints through objective penalties, guidance, or post-processing. CGD instead incorporates  feasibility corrections throughout reverse diffusion, allowing constraint information to shape the generative trajectory before terminal discrete recovery and continuous completion.

\section{Problem Setting and Learning-Assisted Decomposition}

The proposed approach exploits the fact that the combinatorial difficulty of Problem~(\ref{eq:mip}) lies in determining its binary decision variables $z$.
Once these variables are fixed (we write $z = \hat z$), Problem~(\ref{eq:mip}) reduces to the continuous optimization problem
\begin{equation}
\begin{aligned}
\min_{x}\quad f(x,\hat z)\quad
\text{s.t.}\quad & g(x,\hat z)\le 0, \quad 
x\in\mathcal X.
\end{aligned}
\label{eq:continuous_subproblem}
\end{equation}
Although this problem retains the original nonlinearity and nonconvexity, it contains no combinatorial search and can be solved with a standard numerical solver at significantly lower cost.\\
We collect the constraints involving only the binary variables in
\[
\mathcal Z=\{ z\in \{0,1\}^{m} \; | \; h(z;\xi)\le 0\}.
\]
For example, in branch switching, $\mathcal Z$ contains the line configurations satisfying the required connectivity conditions. These constraints can be enforced while generating $z$, independently of the downstream continuous solve.
Given training data $\mathcal D={(\xi_i,z_i^\star)}_{i=1}^{N}$, where $\xi_i$ denotes an instance and $z_i^\star\in\mathcal Z$ is a high-quality reference decision, we learn a conditional generative model that produces a feasible $\hat z\in\mathcal Z$ for an unseen instance $\xi$. The continuous variables are then recovered by solving~(\ref{eq:continuous_subproblem}) with $\hat z$ fixed. This decomposition eliminates the combinatorial search while retaining numerical optimization for continuous completion.

The next section reviews the diffusion framework that forms the basis of our approach and exposes the feasibility gap that motivates the proposed constrained diffusion model.

\section{Diffusion Background and the Feasibility Gap}
 \label{sec:background}
Diffusion models generate samples by learning to reverse a noising process that maps data to a simple prior \citep{song2019generative, song2021score}. 
Let $y_0\sim p_{\mathrm{data}}$ denote a sample in $\mathbb R^d$. 
The forward diffusion process progressively perturbs $y_0$ through a sequence of Markov transitions parameterized by a noise schedule $\{\beta_t\}_{t=1}^{T}$. A key property of this process is that the conditional distribution $q(y_t \mid y_0)$ admits the closed-form representation
$
y_t=\sqrt{\bar\alpha_t}\,y_0+\sqrt{1-\bar\alpha_t}\,\epsilon$, with $\bar\alpha_t=\prod_{i=1}^{t}(1-\beta_i)$ and $ \epsilon\sim\mathcal N(\mathbf 0,\mathbf I),
$
with $y_t$ interpolating between data and noise.

The reverse process is parameterized by a denoising network
$s_\theta(y_t,t,\xi)$. The denoiser may parameterize the reverse process
through a noise estimate or directly through a relaxed estimate of the clean
binary state. In the former case, the denoiser parameters are optimized by minimizing the error between the true noise $\epsilon$ and the predicted noise: 
\begin{equation}
\min_{\theta}
\;
\mathbb E_{t,\,p_{\mathrm{data}}(y_0),\,\mathcal N(\epsilon;\mathbf 0,\mathbf I)}
\left[
\left\|
\epsilon
-s_\theta\big(
y_t, t, \xi \big)
\right\|_2^2
\right]
\label{eq:dsm}
\end{equation}
where $y_t$ is the noisy sample obtained from the forward diffusion process at timestep $t$.  
Sampling starts from $y_T\sim\mathcal N(\mathbf 0,\mathbf I)$ and iterates as
$y_{t-1}
=
\frac{1}{\sqrt{1-\beta_t}}
\left(
y_t
-
\frac{\beta_t}{\sqrt{1-\bar\alpha_t}}
s_\theta(y_t,\xi, t)
\right)
+
\sigma_t \eta,
$ with $
\eta \!\sim\! \mathcal N(\bm 0,\!\bm I)$
, down to $t\!=\!0$, mapping the diffusion prior toward
the data distribution. 
For the constrainedgeneration mechanism developed next, the essential
quantity is the relaxed estimate
$\hat z_t\in[0,1]^m$ of the binary decisions produced at each reverse
step. This representation provides the interface between the diffusion model
and the feasibility-correction operator.

\subsection{Limitations of Diffusion Models for Combinatorial Decisions}
\label{sec:limitations}

Diffusion models can learn distributions over discrete decisions but do not natively enforce application-specific constraints~\citep{Fioretto:NeurIPS24}. This is key because in mixed-integer optimization, even a small discrete error may render the downstream continuous problem infeasible. 
The AC OPF with branch switching provides a representative example
(see Section~\ref{sec:experiments} for details). 
\begin{wrapfigure}[13]{r}{0.4\textwidth}
\vspace{-1.0\baselineskip}
\centering
\includegraphics[width=0.95\linewidth]{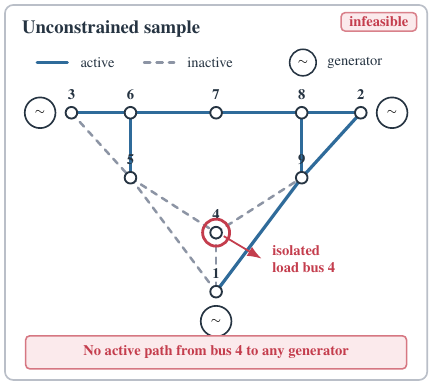}
\captionsetup{font=footnotesize,skip=4pt}
\caption{Infeasible 9-bus topology. Opening (1,4), (4,5), and (4,9) isolates load bus~4.}
\label{fig:infeasible_topology_example}
\end{wrapfigure}
The binary vector $z\in\{0,1\}^m$
specifies active lines, which must connect each load bus to generation.
In Fig.~\ref{fig:infeasible_topology_example}, opening the three lines
incident to bus $4$ isolates its load, making the downstream AC-OPF
infeasible~\citep{PGlib-OPF}. This motivates enforcing combinatorial
feasibility during reverse denoising.
\providecommand{\method}{\textsc{Constrained Graph Diffusion for Combinatorial Decisions}}
\section{Constrained Graph Diffusion for Combinatorial Decisions}
\label{sec:cgd}

\begin{figure}[t]
 \centering
 \vspace{-8pt}
 \includegraphics[width=.95\linewidth, height=6.1cm]{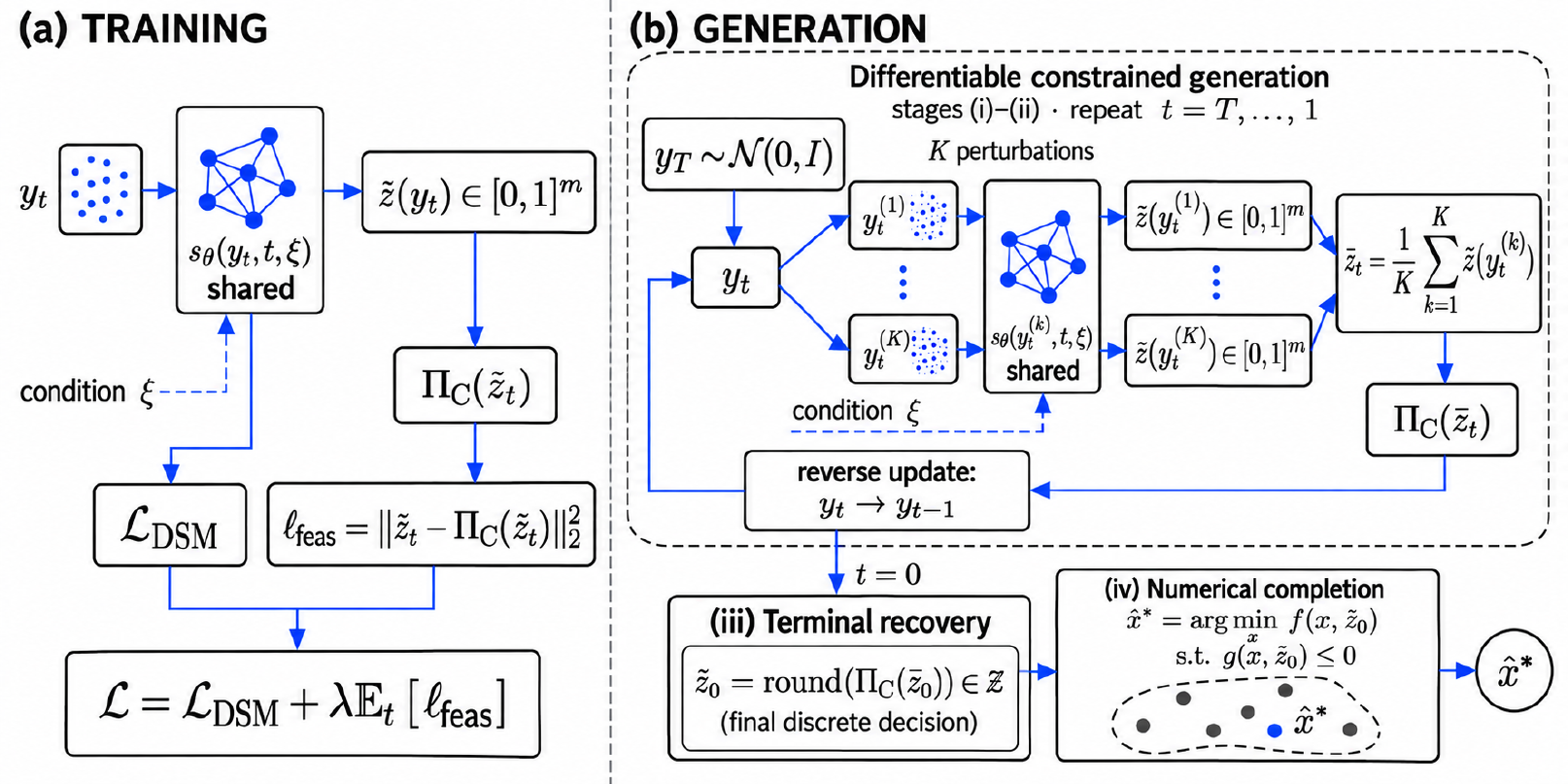}
 \caption{CGD training and generation.
\textbf{(a)} During training, the projected relaxed decision defines the feasibility target used to regularize the diffusion
objective.
\textbf{(b)} During generation, $K$ perturbed predictions are aggregated into a
Monte Carlo estimate used for constraint guidance throughout stages
(i)--(ii), followed by discrete recovery and continuous completion in
stages (iii)--(iv).}
 \label{fig:FGD}
 \vspace{-10pt}
\end{figure}

This section develops Constrained Graph Diffusion (CGD) to generate a high-quality binary decision satisfying the combinatorial requirements of an unseen instance $\xi$. CGD couples a conditional graph diffusion model with a differentiable optimization layer implementing a projection operator. The projection plays two key roles: during training, it defines a feasibility regularizer for the denoiser expected relaxed prediction, as illustrated in Figure~\ref{fig:FGD}(a); during generation, it provides a plug-and-play mechanism that corrects the relaxed decision throughout the diffusion trajectory. 
We defer the complete training procedure to Appendix~\ref{app:training_details} and focus next on
the generation procedure, where constraint information is incorporated throughout the reverse diffusion trajectory.

\textbf{Framework overview.}
At generation time, CGD consists of the following four stages. \emph{(1)} Given an unseen problem instance $\xi$, the graph-based denoiser $s_\theta$ starts from Gaussian noise and estimates a relaxed representation of the binary decision variables. \emph{(2)} At every reverse step, a Monte Carlo estimate of this prediction is projected onto the (relaxed) constrained set and reinserted into the reverse update, steering subsequent predictions toward the application-specific combinatorial constraints. \emph{(3)} After the terminal reverse step, the projected estimate is discretized to obtain the binary decision $\hat z$. \emph{(4)} Finally, $\hat z$ is fixed in the original mixed-integer problem~(\ref{eq:mip}), reducing it to Problem~(\ref{eq:continuous_subproblem}), which is solved with a numerical optimizer to recover the continuous decision $\hat x$. 

\subsection{Constraint-Aware Generation}
\label{sec:constraint_generation}
Given an unseen problem instance $\xi$, the goal of the reverse diffusion process is to generate a feasible binary decision vector $\hat z\in\mathcal Z_\xi$. CGD realizes this goal through a new plug-and-play mechanism for constrained generation, illustrated schematically in Figure~\ref{fig:FGD}(b). This subsection develops stages~(1) and (2) described above, while Section~\ref{sec:recovery} presents stages~(3) and (4).


\textbf{Stage(1): graph-conditioned relaxed decoding.}
The central constraint-enforcement operator in
CGD is a differentiable optimization step (a projection). 
Notice that the constraint-correction operator acts on a relaxed decision in
$[0,1]^m$, whereas the reverse diffusion process evolves in the noisy
continuous state $y_t$. 
Stage (1) therefore maps the denoiser output to a continuous relaxed
representation of the binary decisions on which the projection can act.
A binary decision $z \in \{0,1\}^m$ vector is represented in the diffusion space through 
$e(z) = 2z-1 \in \{-1,1\}$ with $e^{-1}(y)=\frac{y+1}{2},$
which preserves Hamming geometry since
$\|e(z)-e(z')\|_2^2=4d_H(z,z')$.
At reverse timestep $t$, CGD considers $K$ independently perturbed diffusion states $\{ y_t^{(k)}\}$ for the same problem instance $\xi$ and noise level. The denoiser $s_\theta(y^{(k)}_t, t, \xi)$ maps each state to a relaxed estimate of the binary decision, 

\begin{equation}
\hat{z}^{(k)}_t = \hat z_\theta(y^{(k)}_t,t,\xi)
=
\frac{1}{2}
\left[
\tanh\left(
\frac{y^{(k)}_t-\sqrt{1-\bar\alpha_t}\,
s_\theta(y^{(k)}_t,t,\xi)}
{\sqrt{\bar\alpha_t}}
\right)+1
\right]
\in[0,1]^m.
\label{eq:relaxed_prediction}
\end{equation}
These $K$ predictions are aggregated through a Monte Carlo estimate
$
\bar z_t
=
\frac{1}{K}
\sum_{k=1}^{K}
\hat z_t^{(k)},
$
which reduces variance and provides a more stable estimate of the current relaxed decision on which constraint guidance can act~\citep[\S2]{owen2013montecarlo}.

\textbf{Stage(2): Constraint correction.}
Stage (1) produces a relaxed prediction
$\bar z_t\in[0,1]^m$, but this prediction need not satisfy the combinatorial constraints of
instance $\xi$. Stage (2) therefore uses constraint information to correct $\bar z_t$ before reinserting it into the reverse diffusion process.
More formally, given an instance $\xi$, its associated discrete feasible set is $\mathcal Z_\xi
\!=\!
\{z\in\{0,1\}^m:h(z;\xi)\!\leq\!0\}$; direct projection onto $\mathcal Z_\xi$ would require
solving a combinatorial problem at every reverse step. CGD instead uses the continuous relaxation:
\begin{equation}
\qquad
\mathcal C_\xi
=
\{u\in[0,1]^m:\widetilde h(u;\xi)\leq0\}
\qquad
\mathcal C_\xi\cap\{0,1\}^m=\mathcal Z_\xi.
\label{eq:relaxed_feasible_set}
\end{equation}
To project this Monte Carlo prediction $\bar z_t$ onto $C_\xi$, CGD uses the Euclidean projection
\begin{equation}
\Pi_{\mathcal C_\xi}(v)
\in
\arg\min_{u\in\mathcal C_\xi}\|u-v\|_2^2.
\label{eq:projection}
\end{equation}
At timestep $t$, the Monte Carlo prediction $\bar z_t$ is corrected as
$
u_t
=
\Pi_{\mathcal C_\xi}(\bar z_t).
$
The projection leaves a feasible prediction unchanged or makes the smallest Euclidean correction. 
Next, we use the discrepancy between the Monte Carlo prediction and its corrected
counterpart to define a guidance 
\begin{equation}
    \ell_{\mathrm{guide}}(t,\xi)
    =
    \left\|
    \bar z_t-
    \operatorname{sg}[u_t]
    \right\|_2^2 ,
    \label{eq:guidance_loss}
\end{equation}
whose gradient $\nabla_{\hat z_t^{(k)}}\ell_{\mathrm{guide}}$ is used to steer the reverse trajectory toward feasibility. In the above, $\operatorname{sg}$ treats the corrected point as a fixed target; by Proposition~\ref{prop:projection_geometry}, this leaves the exact-projection gradient unchanged, while for approximate corrections it defines a fixed guidance target.
Backpropagating these gradients through the shared denoiser and the $K$ perturbations gives a single guidance direction for the current diffusion state,
\begin{equation}
\nabla_{y_t}\ell_{\mathrm{guide}}
=
\frac{2}{K}
\sum_{k=1}^{K}
\left(
\nabla_{y_t}\hat z_\theta(y_t^{(k)},t,\xi)
\right)^\top
\left(
\bar z_t-u_t
\right),
\label{eq:guidance_grad_state}
\end{equation}
where the derivative includes the dependence of each perturbation
$y_t^{(k)}$ on the current state $y_t$. 
The constrained-guidance direction is then combined with the denoising direction for the reverse update:
\begin{equation}
y_{t-1}
=
\frac{1}{\sqrt{1-\beta_t}}
\left(
y_t
-
\gamma_t \nabla_{y_t}\ell_{\mathrm{guide}}
-
\frac{\beta_t}{\sqrt{1-\bar\alpha_t}}\epsilon_t^C
\right)
+
\sigma_t\eta_t,
\label{eq:corrected_reverse}
\end{equation}
where \(\gamma_t\!>\!0\) controls the strength of the constraint guidance at timestep \(t\). 
Therein, $\epsilon_t^C= \frac{y_t-\sqrt{\bar\alpha_t}\,(2u_t-1)}
{\sqrt{1-\bar\alpha_t}}.$ is the noise estimate corresponding to the corrected relaxed decision $u_t$.

The resulting state $y_{t-1}$ is fed back to the denoiser, allowing the constraint information introduced at timestep $t$ to influence all subsequent reverse steps. 
The following result characterizes the exact projection and motivates its use for both reverse generation and constraint-aware training.

\begin{proposition}[Projection geometry]
\label{prop:projection_geometry}
Let $\mathcal C_\xi$ be nonempty, closed, and convex. Then the projection is unique, the squared distance
$d_{\mathcal C_\xi}^2(v)=\|v-\Pi_{\mathcal C_\xi}(v)\|_2^2$
is continuously differentiable, and
\begin{equation*}
\nabla_v d_{\mathcal C_\xi}^2(v)
=
2\bigl(v-\Pi_{\mathcal C_\xi}(v)\bigr).
\label{eq:projection_gradient}
\end{equation*}
Moreover, for every $z^\star\in\mathcal Z_\xi$,
\begin{equation*}
\|\Pi_{\mathcal C_\xi}(v)-z^\star\|_2^2
\leq
\|v-z^\star\|_2^2
-
\|v-\Pi_{\mathcal C_\xi}(v)\|_2^2.
\label{eq:projection_improvement}
\end{equation*}
\end{proposition}

The proof is provided in Appendix~\ref{prop:projection_geometry_proof}.  Thus, the exact projection is a minimum-distance correction 
whose application cannot increase the distance to any feasible reference, and yields the least disruptive feasible modification of the reverse transition. Let
\begin{equation}
R_t^v(y_t,\eta_t)
=
a_ty_t+b_te(v)+\sigma_t\eta_t,
\qquad
b_t=
\frac{\beta_t\sqrt{\bar\alpha_{t-1}}}
{1-\bar\alpha_t},
\label{eq:endpoint_transition}
\end{equation}
denote the reverse transition anchored at a relaxed endpoint $v$.

\begin{proposition}[Least-disruptive feasible transition]
\label{prop:least_disruptive}
Let $\mathcal C_\xi$ be nonempty, closed, and convex, and let
$u_t=\Pi_{\mathcal C_\xi}(v_t)$. For every $w\in\mathcal C_\xi$ and common noise realization $\eta_t$,
\begin{equation*}
\begin{aligned}
\|R_t^{u_t}-R_t^w\|_2^2
\leq
\|R_t^{v_t}-R_t^w\|_2^2
-
4b_t^2\|v_t-u_t\|_2^2.
\end{aligned}
\label{eq:transition_improvement}
\end{equation*}
\end{proposition}


The proof is provided in Appendix~\ref{prop:least_disruptive_proof}. 
Proposition~\ref{prop:least_disruptive} gives a local guarantee for the least-disruptive feasible reverse transition; because corrected states are fed back to the denoiser, these corrections also influence subsequent predictions.
This trajectory-level effect is reflected empirically in Section~\ref{sec:experiments}, where stepwise correction improves downstream objective quality over unconstrained generation and final-step correction.

\subsection{Discrete Recovery and Continuous Completion}
\label{sec:recovery}

\textbf{Stage (3): discrete recovery. } Note that, after projection, the terminal estimate $u_0$ remains continuous, requiring a discretization operation. This step is deferred until this point because hard decisions made at an earlier timestep would discard the uncertainty needed by the remaining reverse process. A recovery map then converts $u_0$ into a binary decision:
\begin{equation}
\hat z=R_\xi(u_0),
\qquad
R_\xi:[0,1]^m\rightarrow\{0,1\}^m.
\label{eq:recovery}
\end{equation}
The default recovery uses componentwise thresholding,
$
T(u)_i=\mathbf1\{u_i\geq1/2\}.
$
The default recovery uses $R_\xi=T$, but a problem-specific repair can augment this map when available, as in the case of our experiments. Membership of $u_0$ in $C_\xi$ does not alone imply that $T(u_0)$ belongs to $\mathcal Z_\xi$. The following result quantifies what exact convex projection preserves through thresholding.

\begin{corollary}[Threshold-recovery error]
\label{cor:threshold_recovery}
For every $u\in[0,1]^m$ and $z^\star\in\{0,1\}^m$,
\begin{equation}
d_H(T(u),z^\star)
\leq
4\|u-z^\star\|_2^2.
\label{eq:threshold_error}
\end{equation}
If $ C_\xi$ is nonempty, closed and convex, $u=\Pi_{\mathcal C_\xi}(v)$ and
$z^\star\in\mathcal Z_\xi$, then
\begin{equation}
d_H(T(u),z^\star)
\leq
4\left(
\|v-z^\star\|_2^2-\|v-u\|_2^2
\right).
\label{eq:projected_threshold_error}
\end{equation}
\end{corollary}


The proof is provided in Appendix~\ref{cor:threshold_recovery_proof}. Corollary~\ref{cor:threshold_recovery} connects the projection geometry to the Hamming and exact-reconstruction metrics used in Section~\ref{sec:experiments}. It essentially shows that exact projection improves an upper bound on the error to every feasible reference decision. However, the bound certifies discrete feasibility only when it implies exact recovery of such a reference. More generally, a recovery map satisfying $R_\xi(C_\xi)\subseteq\mathcal Z_\xi$ certifies the combinatorial condition; Appendix~\ref{app:rounding_certificates} gives a sufficient slack condition for thresholding to have this property at a particular terminal point.

\textbf{Stage (4): continuous completion.} For a recovered decision $z$, define its continuous completion set
\begin{equation}
\mathcal X_\xi(z)
=
\{x\in\mathcal X:g(x,z;\xi)\leq0\}.
\label{eq:completion_set}
\end{equation}
which is used to recover the continuous decision $\hat x$ by solving Problem~(\ref{eq:continuous_subproblem}) with $z=\hat z$.  The next proposition states when these steps produce a solution that satisfies both the discrete and continuous constraints of the original problem.

\begin{proposition}[Feasibility by composition]
\label{prop:end_to_end}
Suppose that the terminal correction returns $u_0\in\mathcal C_\xi$,
$R_\xi(\mathcal C_\xi)\subseteq\mathcal Z_\xi$, and
$\mathcal X_\xi(z)$ is nonempty for every $z\in\mathcal Z_\xi$. If the completion procedure returns
$\hat x\in\mathcal X_\xi(\hat z)$ whenever this set is nonempty, then
$\hat z=R_\xi(u_0)$ and $\hat x$ form a feasible solution of Problem~\ref{eq:mip}. A globally solved completion is optimal only conditional on $\hat z$.
\end{proposition}

The proof is provided in Appendix~\ref{app:prop_end_to_end}.
Proposition~\ref{prop:end_to_end} shows that a feasible final solution is guaranteed when the corrected prediction satisfies the relaxed constraints, recovery produces a feasible binary decision, and the continuous solution satisfies the remaining constraints. 
For the portfolio problem considered next,
a nonempty support satisfying the quadratic combinatorial constraint admits a continuous completion. 
However, for AC transmission switching, the load-relevant connectivity condition used by CGD is necessary but not sufficient for the nonlinear AC equations; thus the evaluation fixes $z=\hat z$, solves Problem~(\ref{eq:continuous_subproblem}) with an off-the-shelf numerical optimizer, and reports discrete violations, downstream feasibility, objective quality, and end-to-end runtime separately.

\FloatBarrier
\section{Experiments}
\label{sec:experiments}

In this section we evaluate CGD on two hard mixed-integer optimization problems with fundamentally different combinatorial structures: AC optimal power flow with branch switching~\citep{4492805} and portfolio optimization~\citep{rubinstein2002markowitz} with discrete asset selection. The experiments address three questions: \emph{(1) Discrete quality:} does CGD recover high-quality combinatorial decisions? \emph{(2) Feasibility and downstream quality:} do these decisions induce feasible continuous problems with near-optimal objective values? \emph{(3) Computational advantage:} does learning the discrete decisions reduce end-to-end solving time relative to numerically solving the original MIP?

\textbf{Baselines and evaluation.}
We compare CGD against the following baseline methods: MLP and GNN, which directly predict the discrete decisions using feed-forward and graph-based architectures, DIFUSCO~\citep{sun2023difusco}, a SOTA graph-based diffusion framework for combinatorial optimization, and LTO for MIP ~\citep{tang2025learningoptimizemixedintegernonlinear}, which provides a learning-to-optimize baseline for mixed-integer problems by combining learned integer predictions with feasibility correction. Additionally, we perform an ablation study on CGD using two variants: \textbf{CGD without projection}, which removes the proposed feasibility guidance during denoising, and \textbf{CGD + final-step projection}, which applies a projection only after the final denoising step as a post-processing repair. Separately, to evaluate the computational cost, we compare CGD against Gurobi 13.0, the state-of-the-art commercial numerical MIP solver.
Tables~\ref{tab:acopf_results} and \ref{tab:portfolio_results} report Hamming
distance and exact reconstruction to quantify recovery of the reference
discrete decision, while constraint violations, downstream infeasibility, and
objective gap assess downstream performance. We refer to Appendix~\ref{app:experimental_details} for exact metric definitions. For the nonconvex AC-OPF benchmarks, reference solutions are the best feasible solutions returned by the solver, whereas for the portfolio MIQPs we use optimal solutions. For space limitations, additional implementation details and solver settings are reported in Appendix~\ref{app:implementation_details}.

\subsection{AC Optimal Power Flow with Branch Switching}

AC-OPF with branch switching  jointly selects energized lines and generator dispatch while satisfying nonlinear power-flow  constraints. We consider the 9-, 197-, and 500-bus systems~\citep{PGlib-OPF}, containing, respectively, $12, 286$ and $597$ binary variables and  $172, 3916$ and $8651$ continuous variables. CGD guides relaxed line decisions using a load-relevant connectivity condition together with constraint information from the downstream AC-OPF problem. Both these corrections promote load-relevant connected network topologies and steer generation toward satisfying the nonlinear AC constraints, whose downstream feasibility is then evaluated separately. The problem formulation, projection operator, and datasets are given in Appendices~\ref{app:acopf_formulation}-\ref{app:acopf_data} and~\ref{app:connectivity_projection}-\ref{app:opf_lagrangian_projection}.  

\begin{table*}[t]
\centering
\scriptsize
\setlength{\tabcolsep}{2.5pt}
\renewcommand{\arraystretch}{0.95}
\begin{tabular}{@{}llcccc@{}}
\toprule
& & \multicolumn{2}{c}{\textbf{Discrete decision quality}} &
\multicolumn{1}{c}{\textbf{Feasibility}} &
\multicolumn{1}{c}{\textbf{Downstream quality}} \\
\cmidrule(lr){3-4}\cmidrule(lr){5-5}\cmidrule(l){6-6}
\textbf{Case} & \textbf{Method} &
\textbf{Ham. (\%)} $\downarrow$ &
\textbf{Exact (\%)} $\uparrow$ &
\textbf{Infeasible (\%)} $\downarrow$ &
\textbf{Gap (\%)} $\downarrow$ \\
\midrule

\rowcolor{tableours}
\multirow{7}{*}{9-Bus}
& \textbf{CGD (ours)}
& \textbf{1.71} {\scriptsize(0.02)}
& \textbf{69.43} {\scriptsize(4.36)}
& \textbf{0.00} {\scriptsize(0.00)}
& \textbf{0.01} {\scriptsize(0.02)} \\

\rowcolor{tableours2}
& CGD without projection
& 3.86 {\scriptsize(0.14)}
& 55.59 {\scriptsize(6.86)}
& 7.55 {\scriptsize(1.21)}
& 0.55 {\scriptsize(0.05)} \\

\rowcolor{tableours2}
& CGD + final-step projection
& 2.71 {\scriptsize(0.09)}
& 57.51 {\scriptsize(9.12)}
& \textbf{0.00} {\scriptsize(0.00)}
& 0.61 {\scriptsize(0.10)} \\

& MLP
& 11.72 {\scriptsize(1.84)}
& 29.66 {\scriptsize(11.92)}
& \textbf{0.00} {\scriptsize(0.00)}
& 0.87 {\scriptsize(0.17)} \\

& GNN
& 13.71 {\scriptsize(2.46)}
& 9.76 {\scriptsize(1.03)}
& 5.67 {\scriptsize(0.84)}
& 0.28 {\scriptsize(0.03)} \\

& DIFUSCO
& 5.36 {\scriptsize(2.19)}
& 41.03 {\scriptsize(7.59)}
& 15.63 {\scriptsize(9.02)}
& 0.58 {\scriptsize(0.32)} \\

& LTO-for-MIP
& 4.52 {\scriptsize(1.84)}
& 47.36 {\scriptsize(17.12)}
& \textbf{0.00} {\scriptsize(0.00)}
& 0.67 {\scriptsize(0.25)} \\

\midrule

\rowcolor{tableours}
\multirow{7}{*}{197-Bus}
& \textbf{CGD (ours)}
& \textbf{0.17} {\scriptsize(0.03)}
& \textbf{41.20} {\scriptsize(4.96)}
& \textbf{0.00} {\scriptsize(0.00)}
& \textbf{1.76} {\scriptsize(0.23)} \\

\rowcolor{tableours2}
& CGD without projection
& 1.14 {\scriptsize(0.27)}
& 27.17 {\scriptsize(10.81)}
& 4.53 {\scriptsize(1.03)}
& 2.99 {\scriptsize(0.51)} \\

\rowcolor{tableours2}
& CGD + final-step projection
& 2.42 {\scriptsize(0.63)}
& 34.79 {\scriptsize(13.28)}
& \textbf{0.00} {\scriptsize(0.00)}
& 2.32 {\scriptsize(0.31)} \\

& MLP
& 24.55 {\scriptsize(4.72)}
& 0.00 {\scriptsize(0.00)}
& 100.00 {\scriptsize(0.00)}
& n/a \\

& GNN
& 3.47 {\scriptsize(0.52)}
& 25.80 {\scriptsize(8.28)}
& 25.18 {\scriptsize(10.42)}
& 2.34 {\scriptsize(0.51)} \\

& DIFUSCO
& 6.36 {\scriptsize(2.19)}
& 21.03 {\scriptsize(6.59)}
& 7.41 {\scriptsize(2.02)}
& 3.64 {\scriptsize(1.29)} \\

& LTO-for-MIP
& 8.72 {\scriptsize(3.61)}
& 27.12 {\scriptsize(6.19)}
& \textbf{0.00} {\scriptsize(0.00)}
& 2.67 {\scriptsize(1.10)} \\

\midrule

\rowcolor{tableours}
\multirow{7}{*}{500-Bus}
& \textbf{CGD (ours)}
& \textbf{2.05} {\scriptsize(0.31)}
& \textbf{33.24} {\scriptsize(5.75)}
& \textbf{0.00} {\scriptsize(0.00)}
& \textbf{0.20} {\scriptsize(0.04)} \\

\rowcolor{tableours2}
& CGD without projection
& 4.79 {\scriptsize(0.51)}
& 24.70 {\scriptsize(6.73)}
& 3.26 {\scriptsize(0.00)}
& 0.25 {\scriptsize(0.11)} \\

\rowcolor{tableours2}
& CGD + final-step projection
& 5.31 {\scriptsize(0.87)}
& 22.68 {\scriptsize(4.53)}
& \textbf{0.00} {\scriptsize(0.00)}
& 0.41 {\scriptsize(0.10)} \\

& MLP
& 7.49 {\scriptsize(0.89)}
& 0.00 {\scriptsize(0.00)}
& 100.00 {\scriptsize(0.00)}
& n/a \\

& GNN
& 5.24 {\scriptsize(0.59)}
& 15.61 {\scriptsize(3.41)}
& \textbf{0.00} {\scriptsize(0.00)}
& 0.55 {\scriptsize(0.17)} \\

& DIFUSCO
& 10.36 {\scriptsize(15.19)}
& 11.03 {\scriptsize(2.59)}
& 21.18 {\scriptsize(4.73)}
& 0.78 {\scriptsize(0.23)} \\

& LTO-for-MIP
& 6.15 {\scriptsize(2.64)}
& 21.53 {\scriptsize(4.70)}
& \textbf{0.00} {\scriptsize(0.00)}
& 0.74 {\scriptsize(0.12)} \\

\bottomrule
\end{tabular}
\caption{Discrete decision quality and downstream AC-OPF performance on $1000$ test instances. Objective gaps are averaged over instances with a feasible downstream solve; n/a indicates that no such instance exists. Values are reported as mean, with $1$ standard deviation in parentheses.}
\label{tab:acopf_results}
\end{table*}

\begin{figure}[t]
\centering\includegraphics[width=\textwidth]{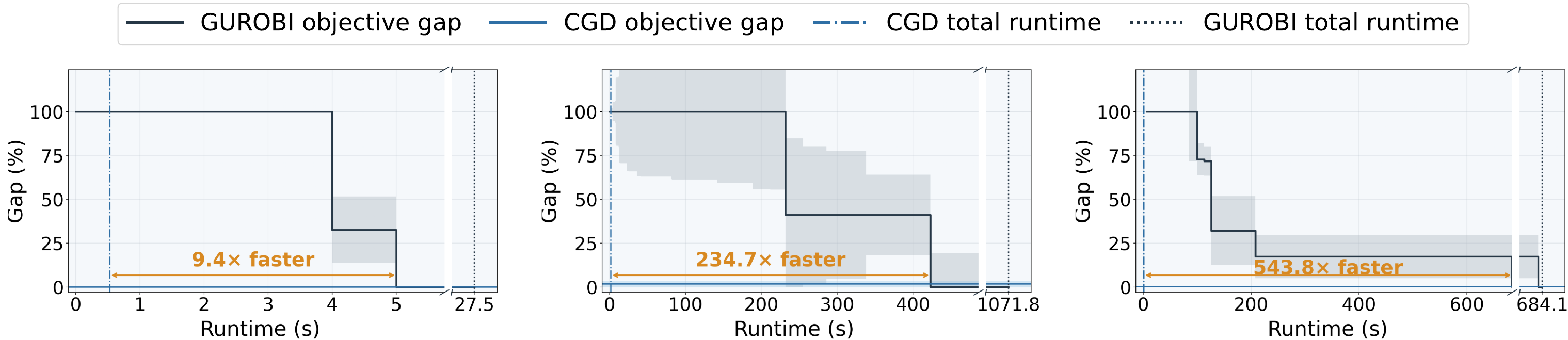}
    \caption{
    Runtime comparison for AC-OPF with branch switching on the
IEEE 9-(left), 197-(middle), and 500-(right) bus systems.
CGD attains objective gaps of $0.01\%$, $1.76\%$, and $0.20\%$ in
$0.53$, $1.80$, and $1.26$\,s, respectively.
The orange arrows report matched-gap speedups of
$9.4\times$, $234.7\times$, and $543.8\times$, computed relative to the
time at which GUROBI first reaches the objective gap achieved by CGD.
Shaded regions indicate one standard deviation across test instances.
    }
    \label{fig:runtime_comparison}
\end{figure}
\textbf{Solution quality and feasibility.} Table~\ref{tab:acopf_results} addresses our first two questions. For \emph{(1) discrete quality}, CGD achieves the lowest Hamming distance and highest exact reconstruction across all three networks, reaching $69.43\%$, $41.20\%$, and $33.24\%$ exact reconstruction on the 9-, 197-,and 500-bus systems, respectively, consistently outperforming the baseline methods. For \emph{(2) feasibility and downstream quality}, CGD achieves zero observed downstream infeasibility on all three networks and the lowest objective gaps of $0.01\%$, $1.76\%$, and $0.20\%$, respectively. 
The ablation shows that the correction mechanism is essential: removing it from generation increases both prediction error and downstream infeasibility, while applying it only at the final step restores feasibility, but does not recover the solution quality of CGD. These results show that incorporating constraint information throughout reverse diffusion improves the generated network topology and downstream performance. 
In addition to these experiments, Appendix~\ref{app:dc_ots} reports results for the DC counterpart (a linear approximation of the AC model, widely used in industry). Interestingly, here, the linear downstream problem enables end-to-end differentiation. The results show the same qualitative pattern than the AC-OPF experiment, with zero observed infeasibility and lower objective gaps than other CGD variants. 
\\[4pt]
\textbf{Runtime.}
Next we investigate question \emph{(3) computational cost}, comparing CGD with Gurobi. For a fair evaluation, Table~\ref{tab:runtime} reports a matched-gap comparison, that is, in each benchmark, we measure the time required by Gurobi to first reach the same objective gap attained
by CGD. We also stress that here, the majority of the time spent by Gurobi is on searching for a good primal solution, as opposed to simply trying to prove the dual bound.
The proposed approach solves the 9-, 197-, and 500-bus instances in $0.53$, $1.80$, and $1.26$ seconds,
yielding increasing speedups with the problem size and complexity, respectively of $9.4\times$, $234.7\times$, and $543.8\times$. 
Figure~\ref{fig:runtime_comparison} visualizes this comparison, illustrating that Gurobi spends practically the majority of the solving time searching for better primal solutions. 
For the linear DC-OPF cases in Appendix~\ref{app:dc_ots}, CGD is slower on the 9-bus system ($0.25\times$), comparable on the 197-bus system ($1.11\times$), and achieves a $16.71\times$ speedup on the 500-bus system. This suggests that the benefit of the proposed decomposition becomes more pronounced as the underlying optimization problem becomes harder.

\subsection{Mixed-Integer Portfolio Optimization}
Next, we also consider a mixed-integer extension of Markowitz portfolio optimization~\citep{rubinstein2002markowitz}, which formulation is reported in Appendix~\ref{app:portfolio_formulation}. Binary variables \(z\) select the assets, while continuous variables \(x\) determine their portfolio weights. The selected support must satisfy the quadratic constraint \(z^\top Pz\le \rho\), after which the weights are recovered through a convex quadratic program. We evaluate CGD and the baseline methods on instances with \(z=50\) and \(z=150\) assets; the correction operator and data-generation algorithm are described in Appendices~\ref{app:portfolio_correction}--\ref{app:portfolio_data}.
\\[2pt]
\textbf{Solution quality and feasibility.}
Table~\ref{tab:portfolio_results} shows that CGD performs strongly in both
\emph{(1)} discrete recovery and \emph{(2)} downstream optimization. For $n=50$, CGD achieves the
lowest Hamming distance ($8.21\%$), highest exact reconstruction ($54.31\%$), and lowest objective gap ($4.92\%$).
For $n=150$, the GNN baseline achieves slightly better reconstruction, yet incurs high constraint violations and a $13.53\%$ objective gap. In contrast, CGD maintains zero observed violation and achieves the lowest objective gap, $6.56\%$. Thus, the accurate reconstruction of the reference support alone is not sufficient for strong downstream performance; constraint-aware generation instead can identify different feasible supports with much better objectives. 
As in AC-OPF, applying the correction throughout generation also outperforms terminal-only correction and unconstrained variants, highlighting the importance of integrating constraints during the reverse diffusion process.
\\[4pt]
\textbf{Runtime.}
Table~\ref{tab:runtime} shows that CGD solves the 50- and 150-asset instances
in $0.020$ and $0.025$ seconds, corresponding to ``matched-gap''
speedups of $1.35\times$ and $17.64\times$ over Gurobi. Similarly to the AC-OPF experiment, the larger speedup on the 150-asset problem shows that CGD becomes more advantageous as the problem size increases. 
\emph{These results are significant: CGD has the potential to enable more frequent and reliable optimization in many large-scale engineering systems.}

\begin{table}[t]
\centering
\scriptsize
\setlength{\tabcolsep}{2.5pt}
\renewcommand{\arraystretch}{0.95}
\begin{tabular}{@{}llccccc@{}}
\toprule
& & \multicolumn{2}{c}{\textbf{Discrete quality}} &
\multicolumn{2}{c}{\textbf{Constraint violation}} &
\multicolumn{1}{c}{\textbf{Downstream quality}} \\
\cmidrule(lr){3-4}\cmidrule(lr){5-6}\cmidrule(l){7-7}
\shortstack{\textbf{Number of}\\\textbf{assets}} &
\textbf{Method} &
\textbf{Ham. (\%)} $\downarrow$ &
\textbf{Exact (\%)} $\uparrow$ &
\textbf{Mean} $\downarrow$ &
\textbf{Max.} $\downarrow$ &
\textbf{Gap (\%)} $\downarrow$ \\
\midrule

\rowcolor{tableours}
\multirow{7}{*}{$50$}
& \textbf{CGD (ours)}
& \textbf{8.21} {\scriptsize(3.91)}
& \textbf{54.31} {\scriptsize(5.46)}
& \textbf{0.00} {\scriptsize(0.00)}
& \textbf{0.00} {\scriptsize(0.00)}
& \textbf{4.92} {\scriptsize(1.71)} \\

\rowcolor{tableours2}
& CGD without projection
& 15.45 {\scriptsize(4.91)}
& 20.32 {\scriptsize(4.38)}
& 0.51 {\scriptsize(0.34)}
& 16.72 {\scriptsize(3.63)}
& 8.43 {\scriptsize(2.95)} \\

\rowcolor{tableours2}
& CGD + final-step projection
& 11.39 {\scriptsize(2.48)}
& 34.76 {\scriptsize(10.32)}
& \textbf{0.00} {\scriptsize(0.00)}
& \textbf{0.00} {\scriptsize(0.00)}
& 8.50 {\scriptsize(3.12)} \\

& MLP
& 27.15 {\scriptsize(5.61)}
& 17.10 {\scriptsize(10.53)}
& 0.03 {\scriptsize(0.01)}
& 15.81 {\scriptsize(3.85)}
& 19.32 {\scriptsize(5.90)} \\

& GNN
& 15.44 {\scriptsize(2.47)}
& 18.20 {\scriptsize(11.51)}
& 0.29 {\scriptsize(0.24)}
& 7.25 {\scriptsize(1.74)}
& 8.82 {\scriptsize(3.15)} \\

& DIFUSCO
& 20.15 {\scriptsize(5.89)}
& 5.90 {\scriptsize(4.17)}
& 0.04 {\scriptsize(0.01)}
& 6.55 {\scriptsize(0.84)}
& 5.79 {\scriptsize(3.12)} \\

& LTO-for-MIP
& 16.91 {\scriptsize(4.63)}
& 0.40 {\scriptsize(0.12)}
& 0.03 {\scriptsize(0.01)}
& 15.80 {\scriptsize(0.78)}
& 19.28 {\scriptsize(6.76)} \\

\midrule

\rowcolor{tableours}
\multirow{7}{*}{$150$}
& \textbf{CGD (ours)}
& 15.84 {\scriptsize(4.12)}
& 44.16 {\scriptsize(13.65)}
& \textbf{0.00} {\scriptsize(0.00)}
& \textbf{0.00} {\scriptsize(0.00)}
& \textbf{6.56} {\scriptsize(2.32)} \\

\rowcolor{tableours2}
& CGD without projection
& 20.29 {\scriptsize(5.45)}
& 29.71 {\scriptsize(9.12)}
& 0.85 {\scriptsize(0.22)}
& 8.14 {\scriptsize(4.76)}
& 11.25 {\scriptsize(5.81)} \\

\rowcolor{tableours2}
& CGD + final-step projection
& 20.26 {\scriptsize(3.12)}
& 28.32 {\scriptsize(3.98)}
& \textbf{0.00} {\scriptsize(0.00)}
& \textbf{0.00} {\scriptsize(0.00)}
& 10.39 {\scriptsize(3.15)} \\

& MLP
& 26.91 {\scriptsize(4.98)}
& 13.09 {\scriptsize(5.22)}
& 0.26 {\scriptsize(0.01)}
& 17.12 {\scriptsize(4.73)}
& 26.30 {\scriptsize(5.62)} \\

& GNN
& \textbf{12.18} {\scriptsize(4.21)}
& \textbf{47.82} {\scriptsize(12.84)}
& 0.48 {\scriptsize(0.12)}
& 5.95 {\scriptsize(1.73)}
& 13.53 {\scriptsize(3.86)} \\

& DIFUSCO
& 18.48 {\scriptsize(5.12)}
& 35.31 {\scriptsize(4.12)}
& 0.36 {\scriptsize(0.18)}
& 5.82 {\scriptsize(2.96)}
& 18.48 {\scriptsize(6.73)} \\

& LTO-for-MIP
& 16.33 {\scriptsize(7.15)}
& 33.43 {\scriptsize(3.70)}
& 0.12 {\scriptsize(0.10)}
& 4.37 {\scriptsize(2.22)}
& 19.23 {\scriptsize(5.13)} \\

\bottomrule
\end{tabular}
\caption{Portfolio decision quality, quadratic-constraint violation, and downstream objective gap on $1200$ test instances. Values are reported as mean, with $1$ standard deviation in parentheses.}
\label{tab:portfolio_results}
\end{table}

\begin{table}[t]
\renewcommand{\arraystretch}{1.1}
\centering
\setlength{\tabcolsep}{6pt}
\resizebox{.7\textwidth}{!}{
\begin{tabular}{@{}lrrrrr@{}}
\toprule
& \multicolumn{3}{c}{\textbf{AC-OPF with branch switching}}
& \multicolumn{2}{c}{\textbf{Portfolio optimization}} \\
\cmidrule(lr){2-4}\cmidrule(l){5-6}
\textbf{Stage}
& \textbf{9-Bus}
& \textbf{197-Bus}
& \textbf{500-Bus}
& \textbf{50 assets}
& \textbf{150 assets} \\
\midrule
\textsc{Gurobi} joint solve
& 27.54 & 1,071.81 & 684.12 & 0.093 & 1.213 \\

\textsc{Gurobi} time to CGD gap
& 4.98 & 422.46 & 684.12 & 0.027 & 0.441 \\

CGD sampling
& 0.13 & 0.16 & 0.11 & 0.014 & 0.015 \\

Continuous completion
& 0.40 & 1.64 & 1.15 & 0.006 & 0.010 \\

\rowcolor{gray!20} \textbf{CGD end-to-end}
& \textbf{0.53}
& \textbf{1.80}
& \textbf{1.26}
& \textbf{0.020}
& \textbf{0.025} \\
\hline
\textbf{Speedup}
& $\mathbf{9.4\times}$
& $\mathbf{234.7\times}$
& $\mathbf{543.8\times}$
& $\mathbf{1.35\times}$
& $\mathbf{17.64\times}$ \\
\bottomrule
\end{tabular}
}
\caption{Average runtime per test instance, in seconds. The speedups are
computed relative to the time required by \textsc{Gurobi} to reach the
objective gap achieved by CGD.}
\label{tab:runtime}
\end{table}

\section{Conclusion}\label{sec: conclude}
This paper presented Constrained Graph Diffusion (CGD), a constraint-aware diffusion framework that formulates discrete generation as a differentiable constrained reverse process over relaxed binary variables. CGD incorporates constraint information directly into the reverse trajectory, steering generation toward the feasible region before terminal recovery. This recovered binary decision is then fixed, and a numerical optimizer recovers the remaining continuous variables. Experiments on AC-OPF with branch switching and mixed-integer portfolio optimization show substantial gains in feasibility and downstream performance with respect to learning-based baselines, while achieving significant speedups over state-of-the-art MIP solvers while sacrificing only a fraction of the optimality gap. Across both applications, these gains increase with problem size, highlighting the importance of separating discrete generation from continuous completion. We believe that CGD can be applied to a wide range of combinatorial optimization problems beyond those studied in this paper, including scheduling, routing, and resource allocation problems.
A possible avenue for future work is the development of foundation diffusion models to allow the resolution of different families of mixed-integer problems and multi-stage problems, where the discrete decisions are made sequentially and the continuous variables are recovered at each stage. 


\section*{Acknowledgements}
This research is partially supported by DOE Genesis project $\#$OE0001051,
NSF awards 2533631, 2401285, 2334936, and by DARPA under Contract No. $\#$HR0011252E005. 
The authors acknowledge the Research Computing at the University of Virginia. Any opinions, findings, conclusions, or recommendations expressed in this material are those of the authors and do not necessarily reflect the views of NSF or DARPA.

\bibliography{iclr2027_conference}
\bibliographystyle{iclr2027_conference}

\appendix
{\Large\textbf{Appendix}}
\section{Additional CGD Details}
\label{app:cgd_details}

This appendix complements Section~\ref{sec:cgd} with parameterization details~\ref{app:relaxed_prediction}--~\ref{app:training_details}, implementation considerations, training derivations, pseudocode (see Algorithm~\ref{alg:cgd}), and proofs~\ref{app:cgd_proofs}.

\subsection{Relaxed Prediction Details}
\label{app:relaxed_prediction}

Under noise prediction, the denoiser first reconstructs the clean embedded state:
\begin{equation}
\widetilde y_{0,t}
=
\frac{
y_t-\sqrt{1-\bar\alpha_t}\,
s_\theta(y_t,t,\xi)
}{
\sqrt{\bar\alpha_t}
},
\qquad
\hat y_{0,t}
=
\tanh(\widetilde y_{0,t}).
\label{eq:clean_estimate}
\end{equation}
The $\tanh$ map bounds the estimate in $[-1,1]^m$, after which the inverse embedding $e^{-1}(y)=(y+1)/2$ produces Equation~\ref{eq:relaxed_prediction}. Each component has the interpretation
\begin{equation}
p_\theta(z_i=1\mid y_t,\xi)
=
\frac{\hat y_{0,t,i}+1}{2},
\qquad
\hat y_{0,t,i}
=
2p_\theta(z_i=1\mid y_t,\xi)-1.
\label{eq:signed_confidence}
\end{equation}

The relaxed prediction can also be viewed as a finite-temperature smoothing:
\begin{equation}
S_\tau(a)
=
\frac{1}{2}
\left[
\tanh\left(\frac{a}{\tau}\right)+1
\right]
=
\frac{1}{1+\exp(-2a/\tau)},
\qquad \tau>0.
\label{eq:smooth_recovery}
\end{equation}
Equation~\ref{eq:relaxed_prediction} uses $\tau=1$. As
$\tau\downarrow0$, $S_\tau(a)$ approaches the hard decision
$\mathbf1\{a\geq0\}$ away from the threshold. For finite \(\tau\), it provides a differentiable path from the feasibility regularizer to the denoiser during training, while hard recovery remains an inference-time terminal operation.

\subsection{Exact and Approximate Corrections}
\label{app:correction_details}

The exact projection in Equation~\ref{eq:projection} provides the geometric reference for CGD. For complex constraint sets, however, solving it at every reverse step may be computationally prohibitive. The approximate operator $\widetilde\Pi_{\mathcal C_\xi}$ may instead be realized through a fixed number of differentiable optimization iterations or through a surrogate gradient estimator.

In the AC-OPF with branch experiments, we use differentiable primal--dual iterations together with a straight-through estimator for evaluating binary branch status. The full application-specific construction is given in Appendix~\ref{app:opf_lagrangian_projection}.

The exact-projection guarantees in Propositions~\ref{prop:projection_geometry} and~\ref{prop:least_disruptive} are not automatically inherited by an approximate operator. In particular, an approximate correction need not return a point in $\mathcal C_\xi$ unless this is established separately.

The noisy state $y_{t-1}$ also need not belong to
$e(\mathcal C_\xi)$. The one-step result in
Proposition~\ref{prop:least_disruptive} does not telescope through the nonlinear denoiser without additional contractivity assumptions. Repeated correction therefore produces a sequence of locally improved reverse transitions; it does not imply that every noisy state is feasible or that CGD samples exactly from the constrained data distribution.

\subsection{Recovery and Completion Details}
\label{app:recovery_details}

Exact projection improves the threshold-recovery bound in
Corollary~\ref{cor:threshold_recovery}, but relaxed feasibility does not generally survive thresholding. Discrete feasibility can be certified when the recovery map satisfies
\[
R_\xi(\mathcal C_\xi)\subseteq\mathcal Z_\xi.
\]
Depending on the application, $R_\xi$ may therefore use simple thresholding, structured rounding, or a problem-specific recovery rule.

For an approximate correction, the projected threshold bound in
Equation~\ref{eq:projected_threshold_error} is not assumed to hold. The quality and feasibility of the recovered binary decision must instead be assessed explicitly.

Proposition~\ref{prop:end_to_end} additionally requires complete recourse: every recovered discrete decision in $\mathcal Z_\xi$ must admit at least one continuous completion. This assumption is application-dependent. Satisfying the constraints exposed to the correction operator may be insufficient for feasibility of the complete continuous model, particularly for nonlinear physical systems. This motivates reporting discrete violations and downstream feasibility as separate metrics.

\subsection{Constraint-Aware Training Details}
\label{app:training_details}

The training algorithm of CGD uses the same relaxed representation introduced in
Section~\ref{sec:cgd}, but does not use the Monte Carlo construction employed
during generation. For each training pair $(\xi,z^\star)$, a timestep $t$ and
noise realization $\epsilon$ are sampled, yielding
\begin{equation}
y_t
=
\sqrt{\bar\alpha_t}\,e(z^\star)
+
\sqrt{1-\bar\alpha_t}\,\epsilon,
\qquad
\epsilon\sim\mathcal N(\mathbf 0,\mathbf I).
\label{eq:training_noisy_state}
\end{equation}
The shared denoiser maps $y_t$ to the relaxed prediction
\begin{equation}
\hat z_t
=
\hat z_\theta(y_t,t,\xi)
\in[0,1]^m,
\label{eq:training_relaxed_prediction}
\end{equation}
using the smooth parameterization described in
Appendix~\ref{app:relaxed_prediction}.

The relaxed prediction is then corrected to define a feasibility target,
\begin{equation}
p_t
=
\operatorname{sg}
\left[
\widetilde\Pi_{\mathcal C_\xi}(\hat z_t)
\right],
\qquad
\ell_{\mathrm{feas}}(t,\xi,z^\star)
=
\|\hat z_t-p_t\|_2^2,
\label{eq:training_feasibility}
\end{equation}
where $\operatorname{sg}$ denotes stop-gradient. Thus, gradients propagate
through the relaxed prediction and the denoising network, while the corrected
point is treated as a fixed target. The resulting gradient is
\begin{equation}
\nabla_\theta \ell_{\mathrm{feas}}
=
2
\left(
\nabla_\theta \hat z_t
\right)^\top
\left(
\hat z_t
-
\widetilde\Pi_{\mathcal C_\xi}(\hat z_t)
\right).
\label{eq:training_feasibility_gradient}
\end{equation}

When
$\widetilde\Pi_{\mathcal C_\xi}=\Pi_{\mathcal C_\xi}$
is the exact Euclidean projection onto a nonempty, closed, and convex set,
Proposition~\ref{prop:projection_geometry} gives
\begin{equation}
\nabla_{\hat z_t}
d_{\mathcal C_\xi}^2(\hat z_t)
=
2
\left(
\hat z_t-\Pi_{\mathcal C_\xi}(\hat z_t)
\right),
\end{equation}
so the stop-gradient expression in
Equation~(\ref{eq:training_feasibility_gradient}) coincides with the gradient
of the squared distance to $\mathcal C_\xi$. The projection Jacobian is not
required, since the optimizing projection point is already accounted for by
the envelope theorem. For an approximate correction, the same expression
defines a feasibility-target surrogate rather than the exact gradient of a
distance function.

The denoiser is trained with the combined objective
\begin{equation}
\mathcal L(\theta)
=
\mathcal L_{\mathrm{DSM}}(\theta)
+
\lambda_{\mathrm{feas}}
\mathbb E_{(\xi,z^\star),t}
\left[
\ell_{\mathrm{feas}}(t,\xi,z^\star)
\right],
\label{eq:training_objective}
\end{equation}
where $\lambda_{\mathrm{feas}}>0$ balances denoising fidelity and relaxed
feasibility. The denoising term preserves fidelity to the reference decision
distribution, while the feasibility term encourages relaxed predictions
toward the supplied combinatorial feasible set. This training-time correction is distinct from the Monte Carlo guidance used
during generation in Section~\ref{sec:constraint_generation}. During training,
the projection defines a fixed feasibility target for learning the denoiser;
during generation, $K$ perturbed predictions are aggregated and their
projection residual is backpropagated with respect to the current diffusion
state to steer the reverse trajectory.

\begin{algorithm}[t]
\caption{Four-stage generation with CGD}
\label{alg:cgd}
\begin{algorithmic}[1]
\Require Instance $\xi$, denoiser $\hat z_\theta$, recovery map $R_\xi$
\State Sample $y_T\sim\mathcal N(0,I)$
\For{$t=T,\ldots,1$}
    \State Construct $K$ perturbed states
    $\{y_t^{(k)}\}_{k=1}^K$ from the current state $y_t$
    \State $\hat z_t^{(k)}
    \leftarrow \hat z_\theta(y_t^{(k)},t,\xi)$,
    \quad $k=1,\ldots,K$
    \State $\bar z_t
    \leftarrow \frac{1}{K}\sum_{k=1}^K \hat z_t^{(k)}$
    \State $u_t
    \leftarrow \widetilde\Pi_{\mathcal C_\xi}(\bar z_t)$
    \State $\ell_{\mathrm{guide}}
    \leftarrow \|\bar z_t-\operatorname{sg}[u_t]\|_2^2$
    \State Compute $\nabla_{y_t}\ell_{\mathrm{guide}}$
    using Equation~\ref{eq:guidance_grad_state}
    \State $\epsilon_t^C
    \leftarrow
    \dfrac{y_t-\sqrt{\bar\alpha_t}(2u_t-1)}
    {\sqrt{1-\bar\alpha_t}}$
    \State Sample $\eta_t\sim\mathcal N(0,I)$
    \State Update $y_{t-1}$
    using Equation~\ref{eq:corrected_reverse}
\EndFor
\State $u_0\leftarrow u_1$
\Comment{Terminal corrected prediction}
\State $\hat z\leftarrow R_\xi(u_0)$
\Comment{Discrete recovery}
\State Fix $z=\hat z$ and use a numerical optimizer to solve
\[
    \min_{x\in\mathcal X} f(x,\hat z)
    \quad\text{subject to}\quad g(x,\hat z;\xi)\le 0.
\]
    \State \Return $(\hat x,\hat z)$
\end{algorithmic}
\end{algorithm}


\subsection{Proofs}
\label{app:cgd_proofs}

\subsubsection{Proof of Proposition~\ref{prop:projection_geometry}}
\label{prop:projection_geometry_proof}
Let $u=\Pi_{\mathcal C_\xi}(v)$. Because $\mathcal C_\xi$ is nonempty, closed, and convex, $u$ exists and is unique. The squared distance to a closed convex set is continuously differentiable, with
\[
\nabla_v d_{\mathcal C_\xi}^2(v)=2(v-u).
\]

The variational characterization of projection gives
\[
\langle v-u,z-u\rangle\leq0
\qquad
\text{for every }z\in\mathcal C_\xi.
\]
For $z^\star\in\mathcal Z_\xi\subseteq\mathcal C_\xi$,
\[
\|v-z^\star\|_2^2
=
\|v-u\|_2^2+\|u-z^\star\|_2^2
+
2\langle v-u,u-z^\star\rangle.
\]
Since the final inner product is nonnegative,
\[
\|u-z^\star\|_2^2
\leq
\|v-z^\star\|_2^2-\|v-u\|_2^2,
\]
which proves the second inequality in Proposition~\ref{prop:projection_geometry}.
\hfill$\square$

\subsubsection{Proof of Proposition~\ref{prop:least_disruptive}}
\label{prop:least_disruptive_proof}

From Equation~\ref{eq:endpoint_transition} and $e(v)=2v-1$,
\[
R_t^v(y_t,\eta_t)-R_t^w(y_t,\eta_t)
=
2b_t(v-w).
\]
Applying the projection inequality from Proposition~\ref{prop:projection_geometry} with
$v=v_t$, $u=u_t$, and $w\in\mathcal C_\xi$ yields
\[
4b_t^2\|u_t-w\|_2^2
\leq
4b_t^2\|v_t-w\|_2^2
-
4b_t^2\|v_t-u_t\|_2^2,
\]
which proves the inequality stated in Proposition~\ref{prop:least_disruptive}.

For fixed $y_t$, $K_t^{v_t}$ and $K_t^w$ are Gaussian distributions with equal covariance and means differing by $2b_t(v_t-w)$. Therefore,
\begin{equation}
W_2^2(K_t^{v_t},K_t^w)
=
4b_t^2\|v_t-w\|_2^2.
\label{eq:wasserstein_distance}
\end{equation}
Since $u_t=\Pi_{\mathcal C_\xi}(v_t)$, minimizing the right-hand side over
$w\in\mathcal C_\xi$ gives
\begin{equation}
u_t
=
\arg\min_{w\in\mathcal C_\xi}
W_2^2\!\left(K_t^{v_t},K_t^w\right).
\label{eq:wasserstein_projection}
\end{equation}
Thus, among reverse transitions with the same covariance and a relaxed
feasible endpoint, the projected transition is the closest to the original
transition in $2$-Wasserstein distance.
\hfill$\square$

\subsubsection{Proof of Corollary~\ref{cor:threshold_recovery}}
\label{cor:threshold_recovery_proof}
Whenever $T(u)_i\neq z_i^\star$,
$|u_i-z_i^\star|\geq1/2$. Hence,
\[
\mathbf1\{T(u)_i\neq z_i^\star\}
\leq
4|u_i-z_i^\star|^2.
\]
Summing over $i$ proves Equation~\ref{eq:threshold_error}. If
$u=\Pi_{\mathcal C_\xi}(v)$, combining this inequality with
with the projection
inequality from Proposition~\ref{prop:projection_geometry} proves
Equation~\ref{eq:projected_threshold_error}.
\hfill$\square$

\subsubsection{Proof of Proposition~\ref{prop:end_to_end}}
\label{app:prop_end_to_end}
Because $u_0\in\mathcal C_\xi$ and
$R_\xi(\mathcal C_\xi)\subseteq\mathcal Z_\xi$, the recovered decision
$\hat z=R_\xi(u_0)$ belongs to $\mathcal Z_\xi$. Complete recourse gives
$\mathcal X_\xi(\hat z)\neq\emptyset$, and the completion procedure returns
$\hat x\in\mathcal X_\xi(\hat z)$. Thus, $(\hat x,\hat z)$ satisfies the discrete and continuous constraints of Problem~\ref{eq:mip}.

A global solution of the completion problem minimizes the objective only over continuous decisions paired with $\hat z$; it does not compare $\hat z$ with other discrete assignments.
\hfill$\square$

\subsection{A Slack Certificate for Threshold Recovery}
\label{app:rounding_certificates}

Corollary~\ref{cor:threshold_recovery} controls discrete error relative to a feasible reference. The next result instead gives a directly checkable sufficient condition for thresholding a relaxed feasible point without violating its discrete constraints.

\begin{proposition}[Feasibility after threshold recovery]
\label{prop:rounding_slack}
Suppose
\[
C_\xi
=
\left\{
u\in[0,1]^m:
\widetilde h_j(u;\xi)\le0,
j=1,\ldots,q
\right\},
\]
where $\widetilde h_j$ agrees with $h_j$ on binary points and is $L_j$-Lipschitz under a fixed norm. For $u\in C_\xi$, let $z=T(u)$ and define
\[
s_j(u):=-\widetilde h_j(u;\xi),
\qquad
\delta(u):=\|T(u)-u\|.
\]
Then, with $[r]_+:=\max\{r,0\}$,
\begin{equation}
\big[h_j(z;\xi)\big]_+
\le
\big[L_j\delta(u)-s_j(u)\big]_+.
\label{eq:rounding_slack_certificate}
\end{equation}
Consequently, $T(u)\in\mathcal Z_\xi$ whenever $L_j\delta(u)\le s_j(u)$ for every $j$.
\end{proposition}

\paragraph{Proof.}
Since $z$ is binary and $\widetilde h_j$ agrees with $h_j$ on binary points, Lipschitz continuity gives
\[
h_j(z;\xi)
=
\widetilde h_j(z;\xi)
\le
\widetilde h_j(u;\xi)
+
L_j\|z-u\|
=
-s_j(u)+L_j\delta(u).
\]
Taking positive parts proves Equation~(\ref{eq:rounding_slack_certificate}).

Proposition~\ref{prop:rounding_slack} identifies the two quantities governing recovery: the relaxed constraint slack $s_j(u)$ and the integrality defect $\delta(u)$. Relaxed feasibility alone requires only $s_j(u)\ge0$; thresholding is certified only when this slack is large enough to absorb the rounding displacement. The condition is sufficient, not necessary, and it applies only when the application admits the stated Lipschitz extension.

\subsection{Portfolio Recovery and Continuous Completion}
\label{app:portfolio_rounding}

\begin{proposition}[Quadratic feasibility after portfolio recovery]
\label{prop:portfolio_rounding}
Let $P=P^\top\succeq0$, let $u\in[0,1]^m$ satisfy $u^\top Pu\le\rho$, and let $z=T(u)$. Define
\[
d:=z-u,
\qquad
r:=\|d\|_2,
\qquad
s:=\rho-u^\top Pu,
\qquad
\lambda:=\lambda_{\max}(P).
\]
Then
\begin{equation}
\big[z^\top Pz-\rho\big]_+
\le
\left[
-s
+
2\|Pu\|_2r
+
\lambda r^2
\right]_+
\le
\left[
-s
+
2\sqrt{\lambda\,u^\top Pu}\,r
+
\lambda r^2
\right]_+.
\label{eq:portfolio_rounding_certificate}
\end{equation}
In particular, thresholding preserves the quadratic constraint whenever $2\|Pu\|_2r+\lambda r^2\le s$.
\end{proposition}
Proposition~(\ref{prop:portfolio_rounding}) characterizes preservation of the quadratic constraint under
componentwise threshold recovery. In the experiments, the recovery map
additionally enforces the cardinality and nonempty-support conditions,
$1\le \mathbf 1^\top z\le k$. These conditions are therefore enforced by
construction, while quadratic feasibility is evaluated separately after
recovery.

\paragraph{Proof.}
Expanding the recovered decision $z=u+d$ gives
\[
z^\top Pz-\rho
=
-s
+
2u^\top Pd
+
d^\top Pd.
\]
Cauchy-Schwarz and positive semidefiniteness give
\[
2u^\top Pd
\le
2\|Pu\|_2\|d\|_2,
\qquad
d^\top Pd
\le
\lambda\|d\|_2^2.
\]
Finally, diagonalizing $P$ gives $\|Pu\|_2^2\le\lambda u^\top Pu$, proving Equation~(\ref{eq:portfolio_rounding_certificate}).

\begin{corollary}[Portfolio completion]
\label{cor:portfolio_completion}
For the fixed-support constraints in Problem~(\ref{eq:portfolio_miqp}), a continuous completion exists if and only if $z\neq\mathbf0$. Consequently, a recovered support induces a feasible portfolio problem whenever $z^\top Pz\le\rho$ and $z\neq\mathbf0$.
\end{corollary}

\paragraph{Proof.}
If $z_i=1$ for some asset $i$, then $x=\mathbf e_i$ satisfies $\mathbf1^\top x=1$ and $0\le x_j\le z_j$ for every $j$. If $z=\mathbf0$, the linking constraints force $x=\mathbf0$, contradicting $\mathbf1^\top x=1$.

Proposition~\ref{prop:portfolio_rounding} explains why exact projection onto the relaxed quadratic set does not alone certify the thresholded support: a projection near the boundary can have insufficient slack $s$ to absorb its integrality defect $r$. Corollary~\ref{cor:portfolio_completion} adds the distinct nonempty-support condition required by the downstream budget equality. These certificates are a posteriori conditions, not additional operations performed by CGD.

\section{Related Works}
\label{app:related_work}
Our work lies at the intersection of learning-for-optimization and diffusion models for combinatorial optimization. Here we provide an overview of the most closely related literature in these two areas.

\textbf{Learning for Mixed-Integer Optimization.}
When many optimization instances share a formulation and differ only in their parameters, learned proxies can amortize computation by mapping instance data to solutions or useful algorithmic decisions~\citep{donti2017task,kotary2021end,park2023self,JMLR:v23:21-0308,bengio2021machine}. Although this paradigm is particularly natural for continuous problems with locally regular solution maps, integrality makes the corresponding map discontinuous and potentially set-valued: small parameter changes can switch the optimal discrete assignment, and multiple structurally different assignments can be near-optimal.

One major line of work therefore learns selected components of a conventional MIP solver. Early learning-to-branch methods imitate strong branching using features of candidate variables~\citep{khalil2016learning}; later work represents a MIP as a bipartite variable--constraint graph and uses graph neural networks to transfer branching policies across instances~\citep{gasse2019exact}. Hybrid architectures reduce the inference overhead of graph-based branching~\citep{gupta2020hybrid}, while Neural Diving and Neural Branching learn primal partial assignments and branching decisions within a solver pipeline~\citep{nair2020solving}. Related methods learn cut selection~\citep{tang2020learningcut,paulus2022learningcut} or graph-based construction heuristics~\citep{khalil2017learning}. For solver-embedded methods, the learned model proposes valid algorithmic actions while the surrounding solver remains responsible for feasibility, bounds, and certification when run to completion. Their scope is nevertheless deliberately local: training often requires expensive expert decisions such as strong-branching or lookahead labels, policy inference adds overhead at many search nodes, and the resulting method still performs online branch-and-bound or solves residual MIPs. Consequently, learned solver components accelerate combinatorial search but do not amortize it away.

A second line learns more direct mappings from parameters to MIP solutions or optimization strategies. Examples include predicting a reusable optimal strategy and reconstructing the associated solution~\citep{Bertsimas_learn_19}, training construction or decision-focused models through continuous surrogates~\citep{wilder2018melding,wilder2019end}, differentiating through a cutting-plane representation of a MIP~\citep{ferbermipaal}, and designing feasibility-aware proxies for mixed-integer nonlinear programs~\citep{tang2025learningoptimizemixedintegernonlinear}. These approaches can amortize a larger fraction of the solve, but typically restrict the strategy class, return a single decision, or rely on relaxations, rounding, repair, or a downstream solver. In particular, a deterministic predictor does not represent alternative near-optimal discrete modes, while interpolation through a continuous relaxation provides no inherent guarantee of integrality or combinatorial feasibility.

CGD occupies a different point in this landscape. It neither learns a branching or cutting policy nor differentiates through the complete mixed-integer solve. Instead, it learns a conditional distribution over the discrete block, 
formulates the generation of the discrete
block as a differentiable constrained reverse process, fixes the recovered
binary assignment, and subsequently solves the remaining continuous problem
with a numerical optimizer.
This removes online branch-and-bound search over the discrete variables and can represent multiple plausible assignments, while retaining a numerical optimizer for the continuous variables. It does not remove continuous nonconvexity or, by itself, provide a global-optimality certificate.

\textbf{Diffusion Models for Combinatorial Optimization.}
Recent work has demonstrated that diffusion models provide an effective generative framework for combinatorial optimization by learning distributions over discrete solution spaces. A seminal contribution in this direction is DIFUSCO~\citep{sun2023difusco}, which introduced graph-based diffusion models for combinatorial optimization problems and demonstrated that iterative denoising can recover high-quality solutions for a variety of graph-structured tasks. Subsequent work has extended this paradigm to unsupervised settings, data-free training, and inference-time adaptation, further demonstrating the flexibility of diffusion-based optimization~\citep{sanokowski2024diffusion, hong2024unsupervised, feng2026cam, lei2025diffuada}.

Despite their success, existing diffusion-based optimization methods typically handle constraints through soft objective formulations or post-processing procedures, rather than explicitly enforcing feasibility throughout the generative process. In mixed-integer optimization, however, the discrete decisions directly affect the feasibility of the continuous variables. This motivates our approach, which integrates the problem combinatorial constraints throughout the generation process, enabling a decomposition in which the discrete decisions are learned by a diffusion model while the continuous variables are subsequently recovered by solving the resulting continuous optimization problem.

\section{AC-OPF with Branch Switching  - Projection onto the Load-Relevant Connectivity Set at Training time}
\label{app:connectivity_projection}

For this problem, the training-time correction
$\Pi_{\mathcal C_\xi}$ projects the relaxed network topology prediction
$\hat z\in[0,1]^{|\mathcal E|}$ onto a continuous relaxation of the
load-relevant connectivity set. The projection is constructed so that its
binary points recover the original connectivity requirement: every connected
component containing nonzero demand must contain at least one generator,
whereas load-free islands are permitted.

Let
\[
d_i =
\begin{cases}
1, & \Re(S_i^d)>\epsilon,\\
0, & \text{otherwise},
\end{cases}
\]
where $\epsilon>0$ is a small threshold used to identify load buses, and let
\[
D=\sum_{i\in\mathcal N} d_i.
\]

The projection introduces relaxed branch-status variables
$z_e\in[0,1]$, directed flow variables $f_{ij,e}\geq0$, and generator
supply variables $s_i\geq0$. Given a relaxed prediction $\hat z$, the
projection is

\begin{align}
\Pi_{\mathcal C_\xi}(\hat z)
\in
\arg\min_{z,f,s}\quad
&
\|z-\hat z\|_2^2
\label{eq:proj_obj}
\\
\text{s.t.}\quad
&
f_{ij,e}\le Dz_e,
\qquad
f_{ji,e}\le Dz_e,
&&
\forall e=(i,j)\in\mathcal E,
\label{eq:proj_cap}
\\
&
s_i=0,
&&
\forall i\notin\mathcal G,
\label{eq:proj_gen}
\\
&
\sum_{e\in\delta^+(i)} f_e
-
\sum_{e\in\delta^-(i)} f_e
=
s_i-d_i,
&&
\forall i\in\mathcal N,
\label{eq:proj_balance}
\\
&
\sum_{i\in\mathcal N}s_i=D,
\label{eq:proj_supply}
\\
&
0\le z_e\le1,
\qquad
f_e\ge0.
\label{eq:proj_domain}
\end{align}

The objective performs the minimum Euclidean correction of the relaxed
prediction $\hat z$, consistently with the projection operator introduced in
Equation~(\ref{eq:projection}). Constraints~(\ref{eq:proj_cap}) couple the
continuous branch-status variables to the auxiliary flow, while
Constraint~(\ref{eq:proj_gen}) restricts flow sources to generator buses.
Constraints~(\ref{eq:proj_balance})--(\ref{eq:proj_supply}) require one unit
of flow to be delivered to every load bus. Consequently, when $z$ is binary,
every connected component containing load must contain at least one generator,
whereas components containing no load may remain disconnected.

The relaxation allows the same connectivity construction to act directly on
the continuous predictions produced by the denoiser. During training, this projection is applied to the relaxed prediction
$\hat z_t$ in Equation~(\ref{eq:training_relaxed_prediction}), providing the
corrected target used by the feasibility regularizer in
Equation~(\ref{eq:training_feasibility}). Since the feasible set defined above is convex,
the projection is a convex quadratic program and corresponds directly to the
ideal Euclidean projection analyzed in Proposition~\ref{prop:projection_geometry}. 

\paragraph{Binary recovery for evaluation.}
For evaluation purposes only, we use the same connectivity formulation with binary branch-status
variables, i.e.,  $z_e\in\{0,1\}$.
Given a binary input $\hat z$, the Euclidean objective
$\|z-\hat z\|_2^2$ is equivalent to the Hamming distance. The resulting MILP
therefore computes the closest binary topology satisfying the load-relevant
connectivity condition.

\section{AC-OPF with Branch Switching - Projection Operator at Generation Time}
\label{app:opf_lagrangian_projection}

For this experiment, CGD uses a differentiable correction based on an augmented-Lagrangian formulation
of the AC power-flow constraints in Model~\ref{model:ots_acopf}.

The sampling-time correction acts jointly on the relaxed topology
$\hat z_t$ and on the continuous AC variables
\begin{equation}
    x=\{S_i^r,V_i,S_{ij}\},
\end{equation}
consistently with Model~\ref{model:ots_acopf}, where
\begin{equation}
    S_i^r=P_i^r+\mathrm{j}Q_i^r,
    \qquad
    V_i=V_i^m e^{\mathrm{j}V_i^a}.
\end{equation}
During the correction, the generator injections $S_i^r$ and voltages $V_i$
are updated as continuous variables, while the branch flows $S_{ij}$ are
computed from the AC branch-flow equations using the current topology and
voltages. These continuous variables are used only to guide the topology
correction. After sampling, the final binary topology $\hat z$ is fixed and
the downstream AC-OPF is solved independently with the numerical optimizer.

Although $\hat z_t\in[0,1]^{|\mathcal L|}$ is continuous, the AC equations
are evaluated using binary branch statuses. We therefore use the hard
decision
\begin{equation}
    T(\hat z_t)_{ij}
    =
    \mathbf 1\{\hat z_{t,ij}\geq 1/2\},
\end{equation}
together with a straight-through estimator: the forward evaluation uses
$T(\hat z_t)$, while gradients with respect to the topology are propagated
through the relaxed prediction $\hat z_t$.

The differentiable AC formulation evaluates the constraints of
Model~\ref{model:ots_acopf}. We denote by
$g_{\mathrm{eq}}(x,\hat z_t;\xi)$ the residuals associated with active- and
reactive-power balance and the reference-bus angle condition, and by
$g_{\mathrm{ineq}}(x,\hat z_t;\xi)\geq0$ the positive violations associated
with connectivity, branch thermal and angle limits, generator active- and
reactive-power limits, and bus-voltage bounds. Hence, satisfaction of the
constraints included in the correction corresponds to
\begin{equation}
    g_{\mathrm{eq}}(x,\hat z_t;\xi)=0,
    \qquad
    g_{\mathrm{ineq}}(x,\hat z_t;\xi)=0.
\end{equation}

To prevent the correction from moving arbitrarily far from the prediction
produced by the denoiser, the augmented Lagrangian additionally contains a
proximity term between the corrected topology $u_t$ and the original relaxed
prediction $\hat z_t$. The sampling-time correction approximately minimizes
\begin{align}
    \mathcal L_\rho
    ={}&
    \lambda_{\mathrm{cost}} f(x,u_t)
    +
    (\lambda^{\mathrm{eq}})^\top
    g_{\mathrm{eq}}(x,u_t;\xi)
    +
    (\lambda^{\mathrm{ineq}})^\top
    g_{\mathrm{ineq}}(x,u_t;\xi)
    \nonumber\\
    &+
    \frac{\rho}{2}
    \left(
        \|g_{\mathrm{eq}}(x,u_t;\xi)\|_2^2
        +
        \|g_{\mathrm{ineq}}(x,u_t;\xi)\|_2^2
    \right)
    +
    \lambda_{\mathrm{anchor}}
    D_{\mathrm{KL}}
    \left(
        u_t\,\|\,\hat z_t
    \right),
    \label{eq:opf_augmented_lagrangian}
\end{align}
where $\lambda^{\mathrm{eq}}$ and
$\lambda^{\mathrm{ineq}}\geq0$ are the equality and inequality multipliers,
respectively, $\rho>0$ is the augmented-Lagrangian penalty parameter,
$\lambda_{\mathrm{anchor}}>0$ controls proximity to the original diffusion
prediction, and $\lambda_{\mathrm{cost}}\geq0$ controls the contribution of
generation cost. Setting $\lambda_{\mathrm{cost}}=0$ yields a
feasibility-only correction.

At each reverse-diffusion steps, a small number of primal-dual iterations
is performed. Starting from $u_t=\hat z_t$, the corrected topology and the
continuous variables are updated by gradient descent on
Equation~(\ref{eq:opf_augmented_lagrangian}),
\begin{align}
    u_t
    &\leftarrow
    u_t-\eta_z\nabla_{u_t}\mathcal L_\rho,
    \\
    x
    &\leftarrow
    x-\eta_x\nabla_x\mathcal L_\rho,
\end{align}
while the dual variables are updated by residual ascent,
\begin{align}
    \lambda^{\mathrm{eq}}
    &\leftarrow
    \lambda^{\mathrm{eq}}
    +
    \eta_\lambda
    g_{\mathrm{eq}}(x,u_t;\xi),
    \\
    \lambda^{\mathrm{ineq}}
    &\leftarrow
    \left[
        \lambda^{\mathrm{ineq}}
        +
        \eta_\lambda
        g_{\mathrm{ineq}}(x,u_t;\xi)
    \right]_+ .
\end{align}
After the prescribed number of inner iterations, the refined topology defines
the sampling-time correction
\begin{equation}
    u_t
    =
    \Pi_{\mathcal C_\xi}(\bar z_t),
\end{equation}
which is reinserted into the reverse diffusion process as described in
Section~\ref{sec:constraint_generation}.

This procedure is an approximate realization of the projection operator
introduced in Section~\ref{sec:constraint_generation}. Unlike the ideal
Euclidean projection analyzed in Propositions~\ref{prop:projection_geometry}--\ref{prop:least_disruptive}, the AC constraint set is
nonlinear and the inner optimization is performed for only a finite number of
iterations. The augmented-Lagrangian terms steer the prediction toward
AC-compatible topologies, while the proximity term retains the
minimum-change principle underlying the ideal projection.

\section{Experimental Details}
\label{app:experimental_details}

\subsection{Baselines}
\label{app:baselines}

We compare CGD against four learning-based baselines and two ablation variants.
\textbf{CGD without projection} uses the same diffusion architecture but
removes constraint correction during training and generation.
\textbf{CGD + final-step projection} applies the correction only after the last
denoising step and represents terminal post-processing.
\textbf{MLP} predicts a relaxed decision in $[0,1]^m$ using a feed-forward
network trained with mean-squared error and thresholds it at $0.5$.
\textbf{GNN} directly predicts the decisions using message passing over the
problem graph.
\textbf{DIFUSCO}~\cite{sun2023difusco} is a graph diffusion model without
application-specific constraint correction during sampling.
\textbf{LTO-for-MIP}~\cite{tang2025learningoptimizemixedintegernonlinear}
predicts integer decisions using differentiable correction layers followed by
feasibility projection.

The MLP and GNN produce one deterministic prediction per instance.
For diffusion-based methods, we use the same Monte Carlo sample budget $K$
during generation. In CGD, these samples correspond to the auxiliary
perturbations used at each reverse step to construct the Monte Carlo estimate
described in Section~\ref{sec:constraint_generation}.

\subsection{Evaluation Metrics}
\label{app:evaluation_metrics}

Let $(x^\star,z^\star)$ denote the reference solution and $\hat z$ the
decision returned by a learned method. For $N$ test instances with $m$ binary
variables, normalized Hamming distance is
\begin{equation}
\label{eq:ham_app}
\mathrm{Ham}\ (\%)
=
\frac{100}{N}
\sum_{j=1}^{N}
\frac{1}{m}
\sum_{i=1}^{m}
\mathbf{1}\!\left[
\hat z_i^{(j)}\neq z_i^{\star(j)}
\right],
\end{equation}
and exact reconstruction is
\begin{equation}
\label{eq:recon_app}
\mathrm{Recon}\ (\%)
=
\frac{100}{N}
\sum_{j=1}^{N}
\mathbf{1}\!\left[
\hat z^{(j)}=z^{\star(j)}
\right].
\end{equation}

We separately report discrete constraint violations and the percentage of
predictions whose induced continuous problem is infeasible. For feasible
instances, the objective gap is
\begin{equation}
\label{eq:obj_gap_app}
\mathrm{Objective\ Gap}\ (\%)
=
\frac{100}{N_{\mathrm{feas}}}
\sum_{j\in\mathcal I_{\mathrm{feas}}}
\frac{
\left|\hat f^{(j)}-f^{\star(j)}\right|
}{
\left|f^{\star(j)}\right|
},
\end{equation}
where
$\hat f^{(j)}=f(\hat x^{(j)},\hat z^{(j)})$,
$f^{\star(j)}=f(x^{\star(j)},z^{\star(j)})$, and
$\mathcal I_{\mathrm{feas}}$ contains the instances with a feasible
continuous completion. The objective gap is therefore conditional on
downstream feasibility.

Throughout the paper, \emph{reference solution} denotes the discrete and
continuous decisions returned by the corresponding optimization solver. For
the nonconvex AC-OPF instances, these solutions are the best feasible solutions
returned under the prescribed time limit and are not assumed to be globally
certified optima.

\begin{table}[h]
\centering
\caption{Scale of the experimental benchmarks.}
\label{tab:benchmark_summary}
\scriptsize
\setlength{\tabcolsep}{6pt}
\begin{tabular}{@{}lrrrr@{}}
\toprule
\rowcolor{tableheader}
\textbf{Benchmark} & \textbf{Binary} & \textbf{Continuous}
& \textbf{Instances} & \textbf{Test} \\
\midrule
IEEE 9-bus   & 12  & 172   & 10,000 & 1,000 \\
IEEE 197-bus & 286 & 3,916 & 10,000 & 1,000 \\
IEEE 500-bus & 597 & 8,651 & 10,000 & 1,000 \\
Portfolio, $n=50$  & 50  & 50  & 12,000 & 1,200 \\
Portfolio, $n=150$ & 150 & 150 & 12,000 & 1,200 \\
\bottomrule
\end{tabular}
\end{table}

\section{AC-OPF with Branch Switching}

\subsection{Problem Formulation}
\label{app:acopf_formulation}

Let $\mathcal N$, $\mathcal L$, and $\mathcal G$ denote the buses,
transmission lines (or branches), and generators. The binary variable $z_{ij}$ indicates
whether line $(i,j)$ is energized. The continuous variables comprise generator
injections $S_i^r$, complex bus voltages $V_i$, and branch flows $S_{ij}$.
The problem is formulated as follows
\begin{model}[H]
\caption{AC-OPF with branch switching.}
\label{model:ots_acopf}
{\small
\begin{subequations}
\begin{align}
\mbox{\bf variables:}\quad
&x=\{S_i^r,V_i,S_{ij}\},\qquad
z_{ij}\in\{0,1\},                                             \nonumber\\
\mbox{\bf minimize:}\quad
&\sum_{i\in\mathcal G}
 c_{2i}\bigl(\Re(S_i^r)\bigr)^2
 +c_{1i}\Re(S_i^r)+c_{0i}                                    \label{top_ac_obj}\\
\mbox{\bf subject to:}\quad
&v_i^l\le |V_i|\le v_i^u,
&& i\in\mathcal N,                                            \label{eq:top_ac_1}\\
&-\theta^\Delta_{ij}z_{ij}
 \le\angle(V_iV_j^*)
 \le\theta^\Delta_{ij}z_{ij},
&& (i,j)\in\mathcal L,                                        \label{eq:top_ac_2}\\
&S_i^{rl}\le S_i^r\le S_i^{ru},
&& i\in\mathcal N,                                            \label{eq:top_ac_3}\\
&|S_{ij}|\le s_{ij}^u z_{ij},
&& (i,j)\in\mathcal L,                                        \label{eq:top_ac_4}\\
&S_i^r-S_i^d=\sum_{(i,j)\in\mathcal L}S_{ij},
&& i\in\mathcal N,                                            \label{eq:top_ac_5}\\
&S_{ij}
=z_{ij}\left(Y_{ij}^*|V_i|^2-Y_{ij}^*V_iV_j^*\right),
&& (i,j)\in\mathcal L,                                        \label{eq:top_ac_6}\\
&\theta_{\mathrm{ref}}=0.                                     \label{eq:top_ac_7}
\end{align}
\end{subequations}
}
\end{model}




\subsection{Data Generation}
\label{app:acopf_data}

For each network, we perturb the parameters
$\xi=\{S^d,Y^*,S^{ru}\}$ and solve the resulting nonconvex mixed-integer
AC-OPF with \textsc{Gurobi}~\cite{gurobi}. For the IEEE 9-bus benchmark, the
active and reactive demands at each load bus share a uniformly sampled scaling
factor within $\pm50\%$ of their nominal values, preserving their correlation.

For the IEEE 197- and 500-bus systems, demands are sampled between $80\%$ and
$120\%$ of nominal values, transmission line parameters between $80\%$ and
$125\%$, and generator upper limits between $90\%$ and $115\%$. Line
resistance parameters are perturbed between $75\%$ and $125\%$, and generator
costs between $80\%$ and $120\%$ of their nominal values.

Each network contains $10{,}000$ instances, split into $80\%$ training,
$10\%$ validation, and $10\%$ test data. A separate model is trained for each
network. Because the underlying formulation is a nonconvex MINLP, reference
solutions are the best feasible solutions returned under the solver settings
and time limit reported in Appendix~\ref{app:implementation_details}. When generating the data, we obtain the following fraction of open lines:
$13.59\%$, $2.49\%$, and $4.70\%$ for AC-OTS on the 9-, 197-, and 500-bus
systems, and $11.20\%$, $3.94\%$, and $7.65\%$ for the corresponding DC-OTS
benchmarks.

\subsection{DC-OPF with Branch Switching}
\label{app:dc_ots}

For completeness, we additionally evaluate CGD on DC-OPF with branch switching, the linearized counterpart of the AC-OPF problem considered
in Section~\ref{sec:experiments}. The DC formulation retains the same combinatorial
line-switching decisions, with binary variables indicating whether transmission
lines are active, but replaces the nonlinear AC power-flow equations with the
DC power-flow approximation. Consequently, once the topology is fixed, the
remaining continuous completion problem is convex and substantially easier to
solve than the corresponding AC-OPF. This provides a complementary setting in
which the combinatorial structure is preserved while the nonconvexity of the
continuous power-flow model is removed. We use this experiment to assess how the
benefit of learning the discrete topology changes when the downstream numerical
optimization problem is comparatively simpler. The resulting prediction quality,
downstream DC-OPF performance, and runtimes are reported in
Tables~\ref{tab:dc_results} and~\ref{tab:dc_runtime}.

During constrained generation, the DC power-flow constraints are incorporated
through a differentiable convex optimization layer~\citep{agrawal2019differentiable}, allowing gradients from the
DC correction to propagate to the relaxed topology decisions. As in the AC
experiments, the recovered topology is fixed and the remaining continuous
DC-OPF is solved numerically.

\begin{model}[H]
\caption{DC-OPF with branch switching.}
\label{model:ots_dcopf}
{\small
\begin{subequations}
\begin{align}
\mbox{\bf variables:}\quad
&x=\{P_i^r,\theta_i,P_{ij}\},\qquad
z_{ij}\in\{0,1\},                                             \nonumber\\
\mbox{\bf minimize:}\quad
&\sum_{i\in\mathcal G}
c_{2i}(P_i^r)^2+c_{1i}P_i^r+c_{0i}
                                                                  \label{top_dc_obj}\\
\mbox{\bf subject to:}\quad
&P_i^{rl}\le P_i^r\le P_i^{ru},
&& i\in\mathcal N,                                             \label{eq:top_dc_1}\\
&-\bar P_{ij} z_{ij}
\le P_{ij}\le
\bar P_{ij} z_{ij},
&& (i,j)\in\mathcal L,                                         \label{eq:top_dc_2}\\
&P_i^r-P_i^d
=
\sum_{(i,j)\in\mathcal L} P_{ij},
&& i\in\mathcal N,                                             \label{eq:top_dc_3}\\
&-M_{ij}(1-z_{ij})
\le
P_{ij}-B_{ij}(\theta_i-\theta_j)
\le
M_{ij}(1-z_{ij}),
&& (i,j)\in\mathcal L,                                         \label{eq:top_dc_4}\\
&\theta_i^l\le\theta_i\le\theta_i^u,
&& i\in\mathcal N,                                             \label{eq:top_dc_5}\\
&\theta_{\mathrm{ref}}=0.                                      \label{eq:top_dc_6}
\end{align}
\end{subequations}
}
\end{model}
Here, $P_i^r$ denotes active-power generation, $\theta_i$ is the voltage
phase angle at bus $i$, and $P_{ij}$ is the active-power flow on line
$(i,j)$. The parameter $B_{ij}$ denotes the line susceptance and
$\bar P_{ij}$ its thermal capacity. Constraint~(\ref{eq:top_dc_4})
enforces the DC power-flow relation when the line is active and relaxes
it through a sufficiently large constant $M_{ij}$ when $z_{ij}=0$.
Constraint~(\ref{eq:top_dc_2}) forces the corresponding power flow to
zero when the line is switched off. Unlike AC-OTS, the DC formulation
removes reactive power, voltage-magnitude variables, and the nonlinear
AC branch-flow equations. With the binary topology fixed, the resulting
continuous completion problem is therefore a convex quadratic program.

\begin{table*}[t]
\centering
\caption{Network topology prediction and downstream DC-OPF with OTS
performance on $N=1{,}000$ test instances per network. Objective gaps are
averaged over instances with a feasible downstream solve; n/a indicates that
no such instance exists. Values are reported as mean (standard deviation).}
\label{tab:dc_results}
\scriptsize
\setlength{\tabcolsep}{3.1pt}
\begin{tabular}{@{}llcccc@{}}
\toprule
\rowcolor{tableheader}
& & \multicolumn{2}{c}{\textbf{Discrete quality}}
& \multicolumn{1}{c}{\textbf{Feasibility}}
& \multicolumn{1}{c}{\textbf{Downstream quality}} \\
\cmidrule(lr){3-4}
\cmidrule(lr){5-5}
\cmidrule(l){6-6}
\rowcolor{tableheader}
\textbf{Case}
& \textbf{Method}
& \textbf{Ham. (\%)} $\downarrow$
& \textbf{Exact (\%)} $\uparrow$
& \textbf{Infeasible (\%)} $\downarrow$
& \textbf{Gap (\%)} $\downarrow$ \\
\midrule

\rowcolor{tableours}
\multirow{7}{*}{9-Bus}
& \textbf{CGD (ours)}
& \textbf{0.43} {\scriptsize(\textbf{0.04})}
& \textbf{58.83} {\scriptsize(\textbf{5.61})}
& \textbf{0.00} {\scriptsize(\textbf{0.00})}
& \textbf{0.25} {\scriptsize(\textbf{0.02})} \\
& CGD without projection
& 3.92 {\scriptsize(0.24)}
& 49.12 {\scriptsize(5.92)}
& 6.18 {\scriptsize(1.21)}
& 1.47 {\scriptsize(0.17)} \\
& CGD + final-step projection
& 1.53 {\scriptsize(0.12)}
& 53.14 {\scriptsize(8.65)}
& 0.00 {\scriptsize(0.00)}
& 1.68 {\scriptsize(0.10)} \\
& MLP
& 8.14 {\scriptsize(1.13)}
& 27.01 {\scriptsize(5.92)}
& 15.74 {\scriptsize(2.01)}
& 2.04 {\scriptsize(0.17)} \\
& GNN
& 10.03 {\scriptsize(0.46)}
& 19.74 {\scriptsize(2.77)}
& 45.23 {\scriptsize(4.19)}
& 2.24 {\scriptsize(0.03)} \\
& DIFUSCO
& 3.61 {\scriptsize(0.53)}
& 35.39 {\scriptsize(6.85)}
& 9.84 {\scriptsize(2.24)}
& 1.51 {\scriptsize(0.32)} \\
& LTO-for-MIP
& 3.81 {\scriptsize(0.75)}
& 32.86 {\scriptsize(5.41)}
& 0.00 {\scriptsize(0.00)}
& 1.43 {\scriptsize(0.25)} \\
\midrule

\rowcolor{tableours}
\multirow{7}{*}{197-Bus}
& \textbf{CGD (ours)}
& \textbf{0.65} {\scriptsize(\textbf{0.05})}
& \textbf{42.59} {\scriptsize(\textbf{5.42})}
& \textbf{0.00} {\scriptsize(\textbf{0.00})}
& \textbf{1.13} {\scriptsize(\textbf{0.17})} \\
& CGD without projection
& 1.21 {\scriptsize(0.38)}
& 32.87 {\scriptsize(6.12)}
& 5.52 {\scriptsize(0.75)}
& 1.32 {\scriptsize(0.29)} \\
& CGD + final-step projection
& 1.25 {\scriptsize(0.31)}
& 34.23 {\scriptsize(5.87)}
& 0.00 {\scriptsize(0.00)}
& 1.98 {\scriptsize(0.21)} \\
& MLP
& 15.32 {\scriptsize(3.54)}
& 0.00 {\scriptsize(0.00)}
& 100.00 {\scriptsize(0.00)}
& n/a \\
& GNN
& 4.53 {\scriptsize(0.77)}
& 28.03 {\scriptsize(4.65)}
& 13.74 {\scriptsize(3.46)}
& 3.34 {\scriptsize(0.23)} \\
& DIFUSCO
& 5.43 {\scriptsize(1.42)}
& 30.67 {\scriptsize(3.90)}
& 10.42 {\scriptsize(1.05)}
& 3.41 {\scriptsize(0.58)} \\
& LTO-for-MIP
& 6.34 {\scriptsize(1.42)}
& 23.91 {\scriptsize(4.48)}
& 0.00 {\scriptsize(0.00)}
& 2.03 {\scriptsize(0.32)} \\
\midrule

\rowcolor{tableours}
\multirow{7}{*}{500-Bus}
& \textbf{CGD (ours)}
& \textbf{1.31} {\scriptsize({0.23})}
& \textbf{41.54} {\scriptsize({4.32})}
& \textbf{0.00} {\scriptsize({0.00})}
& \textbf{0.17} {\scriptsize({0.04})} \\
& CGD without projection
& 3.51 {\scriptsize(0.63)}
& 31.64 {\scriptsize(4.82)}
& 4.85 {\scriptsize(0.60)}
& 1.12 {\scriptsize(0.27)} \\
& CGD + final-step projection
& 4.16 {\scriptsize(0.59)}
& 29.31 {\scriptsize(3.86)}
& 0.00 {\scriptsize(0.00)}
& 1.67 {\scriptsize(0.25)} \\
& MLP
& 8.43 {\scriptsize(1.74)}
& 0.00 {\scriptsize(0.00)}
& 100.00 {\scriptsize(0.00)}
& n/a \\
& GNN
& 6.83 {\scriptsize(0.98)}
& 20.43 {\scriptsize(4.92)}
& 0.00 {\scriptsize(0.00)}
& 1.89 {\scriptsize(0.21)} \\
& DIFUSCO
& 11.23 {\scriptsize(8.25)}
& 8.23 {\scriptsize(1.29)}
& 19.85 {\scriptsize(4.12)}
& 1.12 {\scriptsize(0.23)} \\
& LTO-for-MIP
& 3.87 {\scriptsize(0.44)}
& 38.98 {\scriptsize(5.36)}
& 0.00 {\scriptsize(0.00)}
& 0.98 {\scriptsize(0.25)} \\
\bottomrule
\end{tabular}
\end{table*}

\paragraph{Solution quality and feasibility.}
CGD achieves the lowest Hamming distance and highest exact reconstruction on
all three DC-OPF with branch switching benchmarks while maintaining zero observed downstream
infeasibility. Its objective gaps are $0.25\%$, $1.13\%$, and $0.17\%$ on the
9-, 197-, and 500-bus systems, respectively, consistently improving over both
unconstrained generation and terminal-only correction. In particular,
correction throughout generation reduces the infeasibility observed without
correction from $6.18\%$, $5.52\%$, and $4.85\%$ to zero on the three systems.
These results show that constraint-aware generation remains beneficial even
when the downstream continuous problem is convex.

\paragraph{Runtime.}
We compare CGD with \textsc{Gurobi} at matched objective quality, using the time
required by \textsc{Gurobi} to first reach the objective gap attained by CGD.
On DC-OPF, CGD is slower on the 9-bus system (\(0.25\times\)), roughly comparable on the 197-bus system (\(1.11\times\)), and reaches a \(16.71\times\) matched-gap speedup on the 500-bus system. In contrast, the AC benchmarks exhibit substantially larger speedups, indicating that the proposed decomposition is most beneficial when combinatorial search is coupled with a difficult nonlinear continuous problem.


\begin{table*}[t]
\centering
\caption{Average runtime per DC-OPF test instance, in seconds. Speedup is
computed relative to the time required by \textsc{Gurobi} to reach the
objective gap achieved by CGD.}
\label{tab:dc_runtime}
\scriptsize
\setlength{\tabcolsep}{6pt}
\begin{tabular}{@{}lrrr@{}}
\toprule
\rowcolor{tableheader}
& \multicolumn{3}{c}{\textbf{DC-OPF with branch switching}} \\
\cmidrule(lr){2-4}
\rowcolor{tableheader}
\textbf{Stage}
& \textbf{9-Bus}
& \textbf{197-Bus}
& \textbf{500-Bus} \\
\midrule

\textsc{Gurobi} joint solve
& 0.01 & 0.12 & 18.10 \\

\textsc{Gurobi} time to CGD gap
& 0.01 & 0.10 & 12.87 \\

CGD sampling
& 0.03 & 0.08 & 0.76 \\

Continuous completion
& 0.01 & 0.01 & 0.01 \\

\midrule
\rowcolor{tableours}
\textbf{CGD end-to-end}
& \textbf{0.04}
& \textbf{0.09}
& \textbf{0.77} \\

\textbf{Speedup}
& $\mathbf{0.25\times}$
& $\mathbf{1.11\times}$
& $\mathbf{16.71\times}$ \\

\bottomrule
\end{tabular}
\end{table*}

\section{Portfolio Experimental Details}

\subsection{Mixed-Integer Portfolio Formulation}
\label{app:portfolio_formulation}

We consider the following mixed-integer extension of Markowitz portfolio
optimization~\cite{rubinstein2002markowitz}:
\begin{equation}
\begin{aligned}
\min_{x,z}\quad
&x^\top Qx-\mu^\top x\\
\text{s.t.}\quad
&\mathbf 1^\top x=1,\\
&0\le x_i\le z_i,\qquad i=1,\ldots,n,\\
&z^\top Pz\le\rho,\\
&\mathbf1^\top z\leq k,\\
&z_i\in\{0,1\},\qquad i=1,\ldots,n.
\end{aligned}
\label{eq:portfolio_miqp}
\end{equation}
Here, $\mu$ contains expected returns, $Q$ is the return covariance matrix,
$P\succeq0$ defines a quadratic constraint over the selected assets, and
$\rho$ is its budget. CGD predicts the support $z$ and recovers the weights
$x$ by solving the induced convex quadratic program.

\subsection{Portfolio Correction and Recovery}
\label{app:portfolio_correction}

During reverse diffusion, CGD corrects relaxed selection probabilities toward
\[
\mathcal C_\xi
=
\left\{z\in[0,1]^n | \bm{1}^T z \leq k, z^\top Pz\le\rho\right\}.
\]
Terminal recovery must satisfy both the quadratic selection constraint and the
nonempty-support requirement implied by $\mathbf 1^\top x=1$. Relaxed
projection alone does not generally guarantee feasibility after thresholding.
Appendix~\ref{app:portfolio_rounding} provides a sufficient condition under
which the relaxed quadratic slack absorbs the displacement introduced by
rounding. We therefore report the observed mean and maximum discrete violation
after recovery in addition to downstream objective quality.

\subsection{Portfolio Data Generation}
\label{app:portfolio_data}

Given the problem and data description of~\citet{sambharya2023end}, we construct a mixed-integer extension of a Markowitz portfolio model by introducing binary selection variables, a cardinality upper bound, and a quadratic constraint on the selection vector.
Each entry of the expected-return vector $\mu$ is
sampled independently from $\mathcal U(0,1)$. The quadratic constraint matrix
$P$ is generated to be positive semidefinite, and $\rho$ varies across
instances to produce different feasible asset-selection problems. The
reference MIQP solution supplies the support $z^\star$.

For both $n=50$ and $n=150$, the dataset contains $12{,}000$ instances. Both datasets use an $80\%/10\%/10\%$
training-validation-test split, and a separate model is trained for each
problem size. The average fraction of selected assets is $30.51\%$ for $n=50$ and $12.95\%$ for $n=150$.

\section{Implementation details.}
\label{app:implementation_details}
Both the AC-OPF and DC-OPF branch-switching benchmarks, as well as the
portfolio benchmark, are solved using \textsc{Gurobi}~(v13.0.1). 

For the AC and DC branch-switching benchmarks, the denoiser is an edge-aware GNN
with hidden dimension $256$, $6$ message-passing layers, $2$ conditioning
layers, and a $64$-dimensional timestep embedding. The network is trained with
AdamW using learning rate $2\times10^{-4}$, weight decay $10^{-6}$, and gradient
clipping at $1$. For the AC-OPF benchmark, we use $T=100$ denoising steps,
Monte Carlo sample size $K=20$, and batch size $128$ during training. Downstream AC-OPF
evaluation uses time limits of $60$, $3600$, and $3600\,\mathrm{s}$ for the
9-, 197-, and 500-bus systems, respectively, with a MIP-gap tolerance of
$10^{-4}$.

For the DC-OPF with branch switching experiments, separate models are trained because the DC and AC
formulations produce different reference solutions and network topologies.
The DC models are trained for $100$ epochs using AdamW; batch sizes are $128$,
$64$, and $32$ for the 9-, 197-, and 500-bus systems, respectively.

For the portfolio benchmark, we use $T=30$ denoising steps and Monte Carlo
sample size $K=16$ during generation. The $n=50$ model
uses hidden dimension $160$, $4$ transformer layers, $5$ attention heads,
and batch size $256$, while the $n=150$ model uses hidden dimension $192$,
$4$ transformer layers, $6$ attention heads, and batch size $192$. Both are
trained for up to $300$ epochs with AdamW, learning rate $3\times10^{-4}$,
weight decay $10^{-4}$, dropout $0.1$, and early-stopping patience $50$.

Hyperparameters for all learning-based methods are selected using the
validation set. Training is performed on NVIDIA GPUs, while optimization is
carried out on AMD EPYC/Intel Xeon CPU nodes.

\end{document}